\documentclass[10pt,twocolumn,letterpaper]{article}

\usepackage{cvpr}
\usepackage{times}
\usepackage{epsfig}
\usepackage{graphicx}
\usepackage{amsmath}
\usepackage{amssymb}

\usepackage{enumerate}
\usepackage{booktabs}
\usepackage{algorithm}  
\usepackage{algpseudocode}
\usepackage{multirow}
\usepackage{mathtools}
\usepackage{mathrsfs}
\usepackage{makecell}
\usepackage{caption}
\usepackage{color,xcolor}
\usepackage{subfig}
\usepackage{colortbl}
\usepackage{bm}

\definecolor{grayDark}{gray}{0.95}
\definecolor{grayLight}{gray}{0.98}
\usepackage[pagebackref=true,breaklinks=true,letterpaper=true,colorlinks,bookmarks=false, hypertexnames=false]{hyperref}

\newcommand{\parent}[0] {{\mathit{parent}}}
\newcommand{\bb}[0] {{\bf b}}

\cvprfinalcopy % *** Uncomment this line for the final submission

\def\cvprPaperID{589} % *** Enter the CVPR Paper ID here

\begin{document}

%%%%%%%%% TITLE
\title{Cascaded Deep Monocular 3D Human Pose Estimation with Evolutionary Training Data}
\author{Shichao Li$^{1}$, Lei Ke$^{1}$, Kevin Pratama$^{1}$, Yu-Wing Tai$^{2}$, Chi-Keung Tang$^{1}$, Kwang-Ting Cheng$^{1}$\\	
	$^{1}$The Hong Kong University of Science and Technology, $^{2}$Tencent\\
	{\tt\small slicd@cse.ust.hk},
	\quad{\tt\small timcheng@ust.hk}
}

\maketitle

%%%%%%%%% ABSTRACT
\begin{abstract}
	End-to-end deep representation learning has achieved remarkable accuracy for monocular 3D human pose estimation, yet these models may fail for unseen poses with limited and fixed training data. This paper proposes a novel data augmentation method that: (1) is scalable for synthesizing massive amount of training data (over 8 million valid 3D human poses with corresponding 2D projections) for training 2D-to-3D networks, (2) can effectively reduce dataset bias. Our method evolves a limited dataset to synthesize unseen 3D human skeletons based on a hierarchical human representation and heuristics inspired by prior knowledge. Extensive experiments show that our approach not only achieves state-of-the-art accuracy on the largest public benchmark, but also generalizes significantly better to unseen and rare poses. Code, pre-trained models and tools are available at this HTTPS URL\footnote{https://github.com/Nicholasli1995/EvoSkeleton}.
\end{abstract}

%%%%%%%%% BODY TEXT
\section{Introduction}
Estimating 3D human pose from RGB images is critical for applications such as action recognition~\cite{Luvizon_2018_CVPR} and human-computer interaction, yet it is challenging due to lack of depth information and large variation in human poses, camera viewpoints and appearances. Since the introduction of large-scale motion capture (MC) datasets \cite{sigal2010humaneva, ionescu2013human3}, learning-based methods and especially deep representation learning have gained increasing momentum in 3D pose estimation. Thanks to their representation learning power, deep models have achieved unprecedented high accuracy~\cite{pavlakos2017coarse, nie2017monocular, lin2017recurrent, martinez2017simple, Luvizon_2018_CVPR, sun_2018_eccv}.

\begin{figure}[h]
	\begin{center}
		\includegraphics[width=0.9\linewidth, trim=0.5cm 2cm 0.5cm 0.5cm]{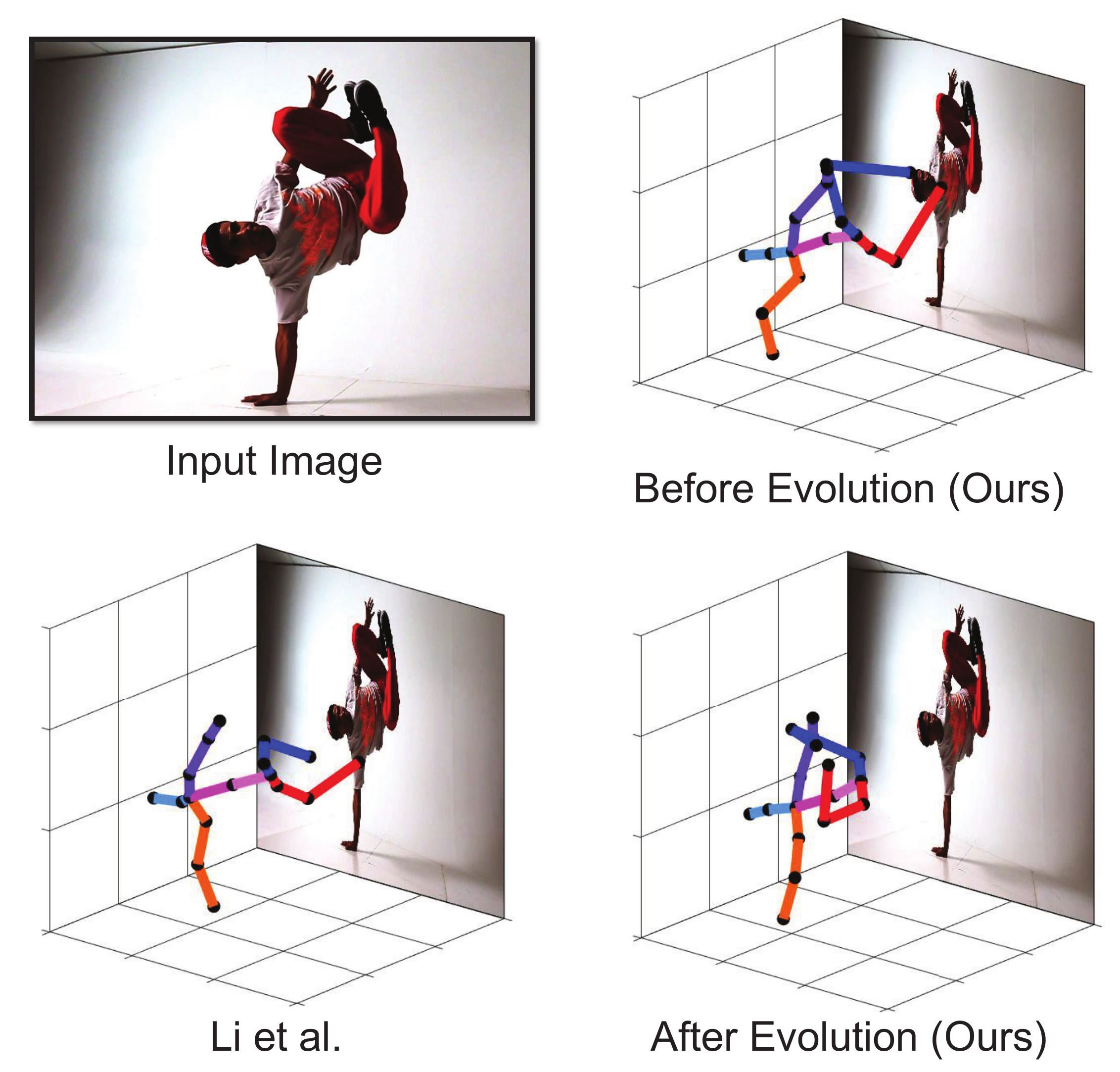}
	\end{center}
	\caption{
		Model trained on the evolved training data generalizes better than~\cite{Li_2019_CVPR} to unseen inputs.}
	\label{fig:teaser}
\end{figure}

Despite their success, deep models are data-hungry and vulnerable to the limitation of data collection. This problem is more severe for 3D pose estimation due to two factors. First, collecting accurate 3D pose annotation for RGB images is expensive and time-consuming. Second, the collected training data is usually biased towards indoor environment and selected daily actions. Deep models can easily exploit these bias but fail for unseen cases in unconstrained environments. This fact has been validated by recent works \cite{zhou2017towards, yang20183d, Li_2019_CVPR, Wandt_2019_CVPR} where cross-dataset inference demonstrated poor generalization of models trained with biased data.

To cope with the domain shift of appearance for 3D pose estimation, recent state-of-the-art (SOTA) deep models adopt the two-stage architecture~\cite{Zhao_2019_CVPR, Cheng_2019_ICCV, Ci_2019_ICCV}. The first stage locates 2D human key-points from appearance information, while the second stage lifts the 2D joints into 3D skeleton employing geometric information. Since 2D pose annotations are easier to obtain, extra in-the-wild images can be used to train the first stage model, which effectively reduces the bias towards indoor images during data collection. However, the second stage 2D-to-3D model can still be negatively influenced by geometric data bias, yet not studied before. We focus on this problem in this work and our research questions are: \textit{are our 2D-to-3D deep networks influenced by data bias? If yes, how can we improve network generalization when the training data is limited in scale or variation?}    

To answer these questions, we propose to analyze the training data with a hierarchical human model and represent human posture as collection of local bone orientations. We then propose a novel dataset evolution framework to cope with the limitation of training data. Without any extra annotation, we define evolutionary operators such as \emph{crossover} and \emph{mutation} to discover novel valid 3D skeletons in tree-structured data space guided by simple prior knowledge. These synthetic skeletons are projected to 2D and form 2D-3D pairs to augment the data used for training 2D-to-3D networks. With an augmented training dataset after evolution, we propose a cascaded model achieving state-of-the-art accuracy under various evaluation settings. Finally, we release a new dataset for unconstrained human pose in-the-wild. Our contributions are summarized as follows:     

\begin{itemize}
	\item To our best knowledge, we are the first to improve 2D-to-3D network training with synthetic paired supervision.
	\item We propose a novel data evolution strategy which can augments an existing dataset by exploring 3D human pose space without intensive collection of extra data. This approach is scalable to produce 2D-3D pairs in the order of 10$^7$, leading to better model generalization of 2D-to-3D networks. 
	\item We present TAG-Net, a deep architecture consisting of an accurate 2D joint detector and a novel cascaded 2D-to-3D network. It out-performs previous monocular models on the largest 3D human pose estimation benchmark in various aspects.
	\item We release a new labeled dataset for unconstrained human pose estimation in-the-wild.
\end{itemize}

Fig.~\ref{fig:teaser} shows a 2D-to-3D network trained on our augmented dataset can handle rare poses while others such as~\cite{Li_2019_CVPR} may fail.

\section {Related Works}
\noindent \textbf{Monocular 3D human pose estimation.}
Single-image 3D pose estimation methods are conventionally categorized into \emph{generative methods} and \emph{discriminative methods}. Generative methods fit parametrized models to image observations for 3D pose estimation. These approaches represent humans by PCA models~\cite{akhter2015pose, Zhou_2016_CVPR}, graphical models~\cite{burenius20133d, belagiannis20143d} or deformable meshes~\cite{anguelov2005scape, loper2015smpl, bogo2016keep, SMPL-X:2019, Kolotouros_2019_ICCV}. The fitting process amounts to non-linear optimization, which requires good initialization and refines the solution iteratively. Discriminative methods~\cite{rosales2002learning, agarwal2005recovering, bo2009structured} directly learn a mapping from image observations to 3D poses. Recent deep neural networks (DNNs) fall into this category and employ two mainstream architectures: \emph{one-stage} methods~\cite{yang20183d, zhou2017towards, Luvizon_2018_CVPR, pavlakos2017coarse, nie2017monocular, lin2017recurrent, sun_2018_eccv, habibie2019wild} and \emph{two-stage} methods~\cite{moreno20173d, martinez2017simple, Hossain_2018_ECCV, Zhao_2019_CVPR}. The former directly map pixel intensities to 3D poses, while the latter first extract intermediate representation such as 2D key-points and then lift them to 3D poses. 

We adopt the discriminative approach and focus on the 2D-to-3D lifting network. Instead of using a fixed training dataset, we evolve the training data to improve the performance of the 2D-to-3D network.

\noindent \textbf{Weakly-supervised 3D human pose estimation.} Supervised training of DNNs demands massive data while 3D annotation is difficult. To address this problem, weakly-supervised methods explore other potential supervision to improve network performance when only few training data is available~\cite{Pavlakos_2017_CVPR, Rhodin_2018_ECCV, Rhodin_2018_CVPR, kocabas2019self, chen2019weakly, wandt2019repnet, Li_2019_ICCV}. Multi-view consistency~\cite{Pavlakos_2017_CVPR, Rhodin_2018_ECCV, Rhodin_2018_CVPR, kocabas2019self, chen2019weakly} is proposed and validated as useful supervisory signal when training data is scarce, yet a minimum of two views are needed. In contrast, we focus on effective utilization of scarce training data by synthesizing new data from existing ones and \emph{uses only single view}.

\noindent \textbf{Data augmentation for pose estimation.} New images can be synthesized to augment indoor training dataset~\cite{rogez2016mocap, varol2017learning}. In~\cite{varol2017learning} new images were rendered using MC data and human models. Domain adaption was performed in~\cite{chen2016synthesizing} during training with synthetic images. Adversarial rotation and scaling were used in ~\cite{peng2018jointly} to augment data for 2D pose estimation. These works produce synthetic images while we focus on data augmentation for 2D-to-3D networks and produce synthetic 2D-3D pairs.   

\noindent \textbf{Pose estimation dataset.} Most large-scale human pose estimation datasets~\cite{zhang2013actemes, lin2014microsoft, andriluka14cvpr} only provide 2D pose annotations. Accurate 3D annotations~\cite{ionescu2013human3, sigal2010humaneva} require MC devices and these datasets are biased due to the limitation of data collection process. Deep models are prone to overfit to these biased dataset~\cite{tommasi2017deeper, torralba2011unbiased, li2019repair}, failing to generalize in unseen situations. Our method can synthesize {\em for free without human annotation} large amount of valid 3D poses with larger coverage in human pose space.
%------------------------------------------------------------------------
\section{Dataset Evolution}
\label{dataevo}
From a given input image $\mathbf{x}_i$ containing one human subject, we aim to infer the 3D human pose $\hat{\mathbf{p}}_i$ given the image observation $\phi(\mathbf{x}_i)$. To encode geometric information as other 2D-to-3D approaches~\cite{martinez2017simple, Zhao_2019_CVPR, Li_2019_CVPR}, we represent $\phi(\mathbf{x})$ as the 2D coordinates of $k$ human key-points $(x_i, y_i)^{k}_{i=1}$ on the image plane. As a discriminative approach, we seek a regression function $\mathcal{F}$ parametrized by $\mathbf{\Theta}$ that outputs 3D pose as $\hat{\mathbf{p}}_i = \mathcal{F}(\phi(\mathbf{x}_i), \mathbf{\Theta})$. This regression function is implemented as a DNN. Conventionally this DNN is trained on a dataset collected by MC devices~\cite{sigal2010humaneva, ionescu2013human3}. This dataset consists of paired images and 3D pose ground truths $\{(\mathbf{x}_{i}, \mathbf{p}_i)\}_{i=1}^{N}$ and the DNN can be trained by gradient descent based on a loss function defined over the training dataset 
$\mathcal{L} = \sum_{i=1}^{N}E(\mathbf{p}_i, \hat{\mathbf{p}}_i)$
where $E$ is the error measurement between the ground truth $\mathbf{p}_i$ and the prediction $\hat{\mathbf{p}}_i=\mathcal{F}(\phi(\mathbf{x}_i), \mathbf{\Theta})$.

Unfortunately, sampling bias exists during the data collection and limits the variation of the training data. Human 3.6M (H36M)~\cite{ionescu2013human3}, the largest MC dataset, only contains 11 subjects performing 15 actions under 4 viewpoints, leading to insufficient coverage of the training 2D-3D pairs $(\phi(\mathbf{x}_i), \mathbf{p}_i)$. A DNN can overfit to the dataset bias and become less robust to unseen $\phi(\mathbf{x})$. For example, when a subject starts street dancing, the DNN may fail since it is only trained on daily activities such as sitting and walking. This problem is even exacerbated for the weakly-supervised methods~\cite{Pavlakos_2017_CVPR, Rhodin_2018_CVPR, chen2019weakly} where a minute quantity of training data is used to simulate the difficulty of data collection.

\begin{figure}[h]
	\begin{center}
		\includegraphics[width=1\linewidth, trim=0.5cm 1.5cm 0.5cm 0.5cm]{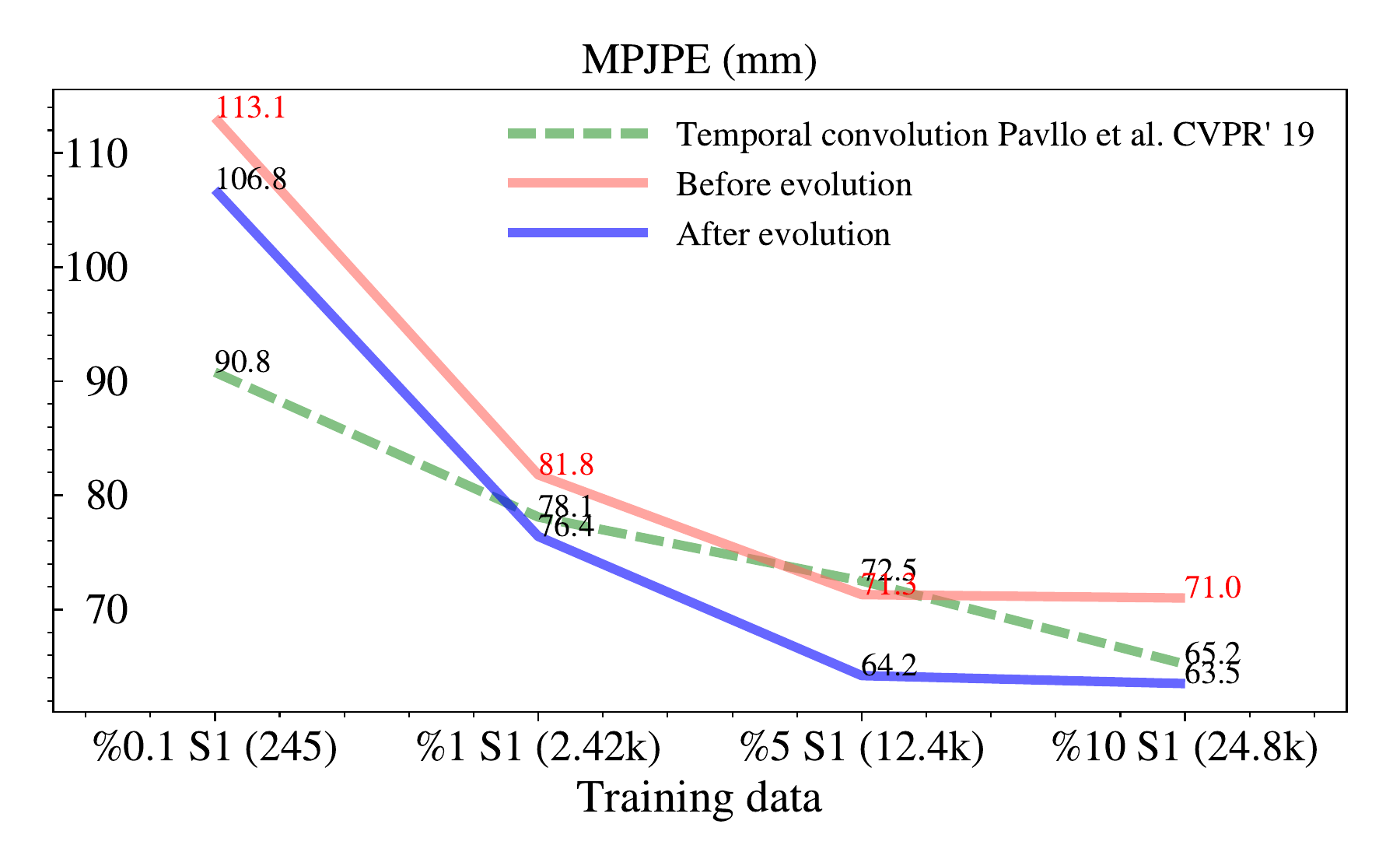}
	\end{center}
	\caption{Generalizing errors (MPJPE using ground truth 2D key-points as inputs) on H36M before and after dataset evolution with varying size of initial population.}
	\label{wsr}
\end{figure}

We take a non-stationary view toward the training data to cope with this problem. While conventionally the collected training data is fixed and the trained DNN is not modified during its deployment, here we assume the data and model can evolve during their life-time. Specifically, we synthesize novel 2D-3D pairs based on an initial training dataset and add them into the original dataset to form the evolved dataset. We then re-train the model with the evolved dataset. As shown in Fig.~\ref{wsr}, model re-trained on the evolved dataset has consistently lower generalization error, comparing to a model trained on the initial dataset. 

In the following we show that by using a hierarchical representation of human skeleton, the synthesis of novel 2D-3D pairs can be achieved by evolutionary operators and camera projection.  

\subsection{Hierarchical Human Representation} \label{hmr}
\begin{figure}
	\begin{center}
		\includegraphics[width=0.9\linewidth, trim=3cm 3cm 0.5cm 0.5cm]{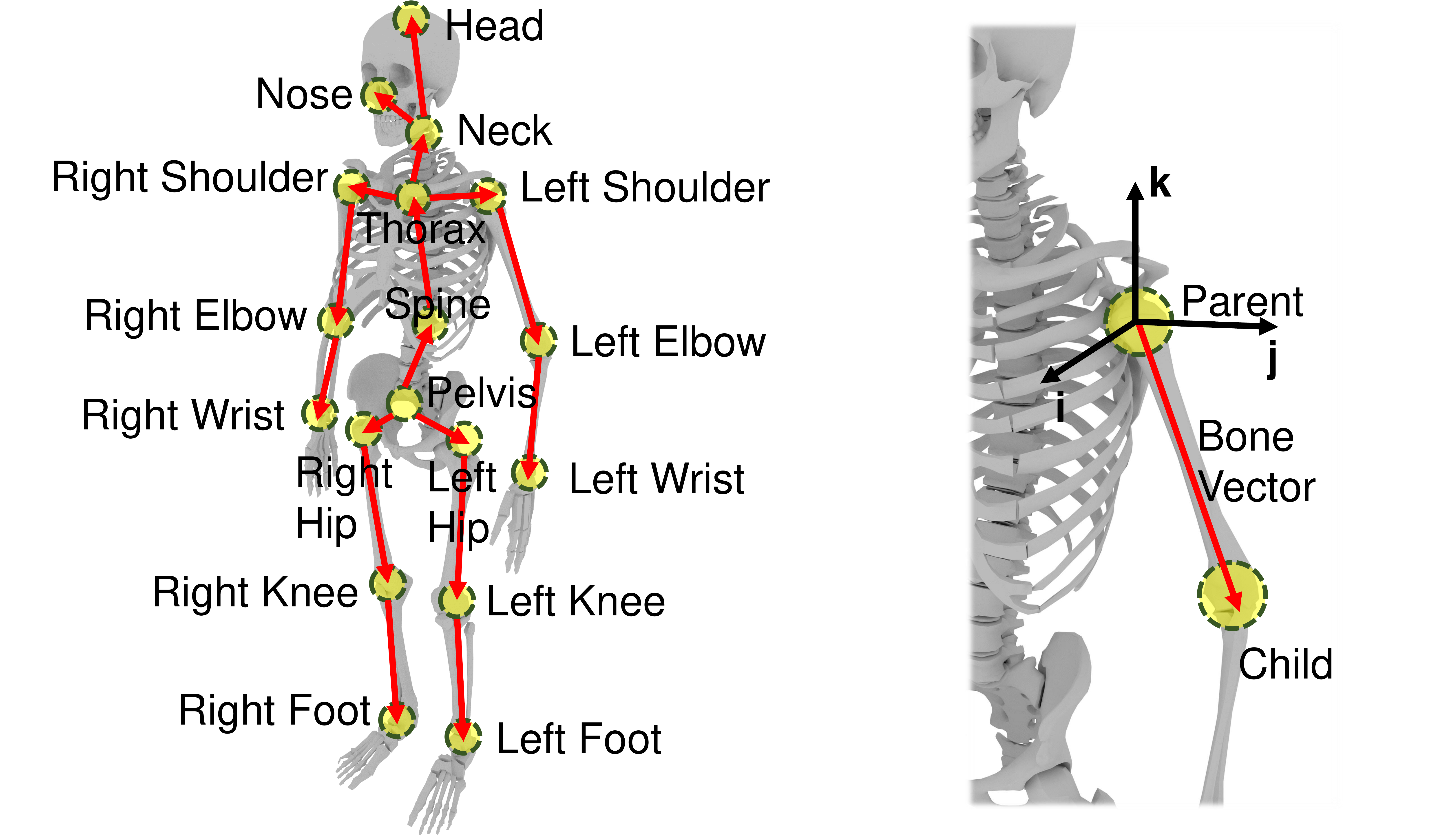}
	\end{center}
	\caption{Hierarchical human representation. Left: 3D key-points organized in a kinematic tree where red arrows point from parent joints to children joints. Right: Zoom-in view of a local coordinate system.}
	\label{skeleton}
\end{figure}

We represent a 3D human skeleton by a set of bones organized hierarchically in a kinematic tree as shown in Fig.~\ref{skeleton}. This representation captures the dependence of adjacent joints with tree edges.

\begin{figure}
	\begin{center}
		\includegraphics[width=0.9\linewidth, trim=3cm 3cm 3cm 2cm]{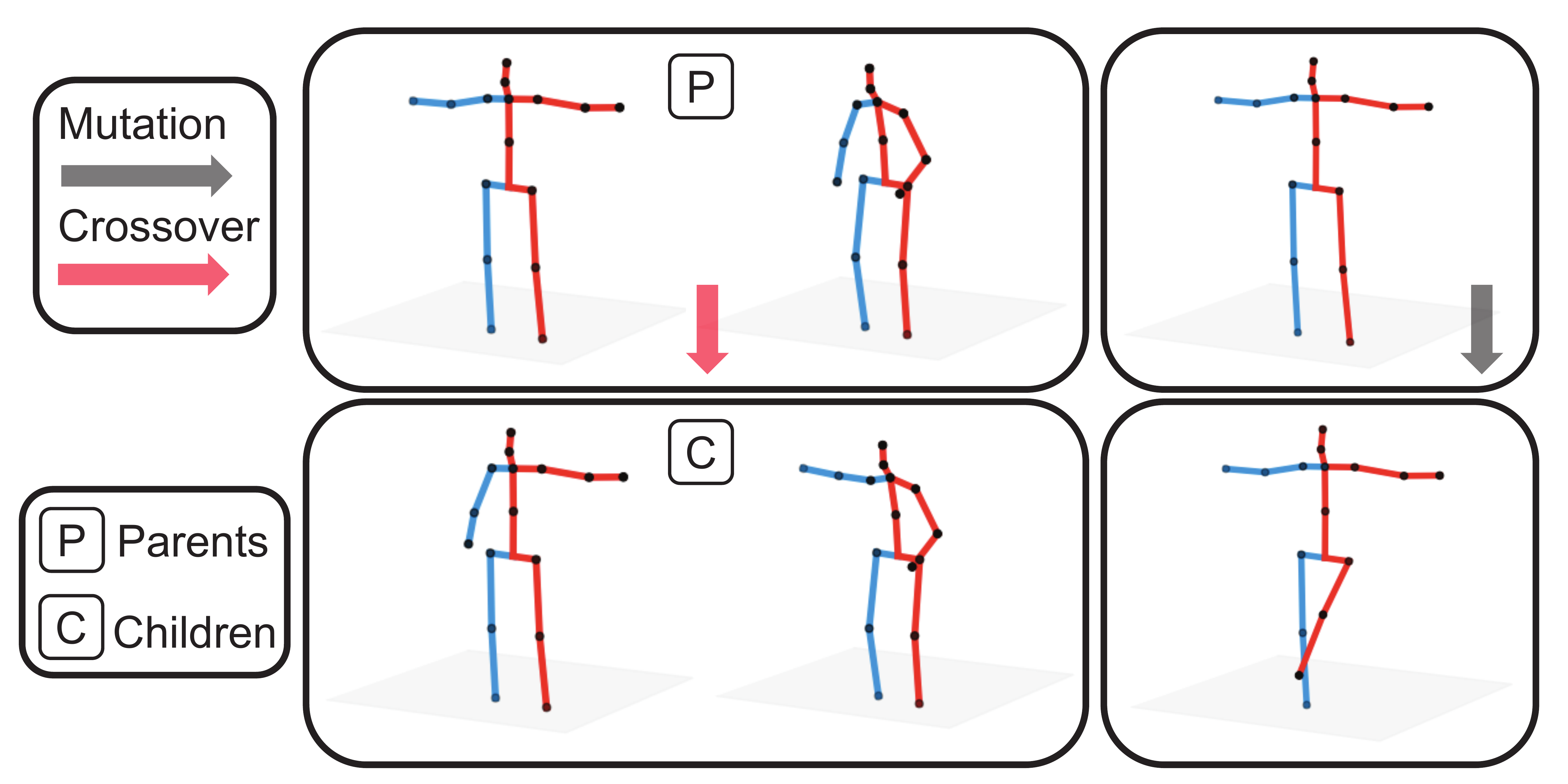}
	\end{center}
	\caption{Examples of evolutionary operation. Crossover and mutation take two and one random samples respectively to synthesize novel human skeletons. In this example the right arms are selected for crossover while the left leg is mutated.}
	\label{fig:data generation}
\end{figure}

Each 3D pose $\mathbf{p}$ corresponds to a set of bone vectors $\{\bb^1, \bb^2, \cdots, \bb^w\}$ and a bone vector is defined as 
\begin{equation}
\bb^{i} = \mathbf{p}^{child(i)} - \mathbf{p}^{parent(i)}
\end{equation}
where $\mathbf{p}^j$ is the $j$th joint in the 3D skeleton and $\parent(i)$ gives the parent joint index of the $i$th bone vector. A local coordinate system\footnote{The coordinate system is detailed in our supplementary material.} is attached at each parent node. For a parent node $\mathbf{p}^{parent(i)}$, its local coordinate system is represented by the rotation matrix defined by three basis vectors $\mathbf{R}^i = [\mathbf{i}^i, \mathbf{j}^i, \mathbf{k}^i]$. The global bone vector is transformed into this local coordinate system as 
\begin{equation}
\label{transformation}
\mathbf{b}^{i}_{local} = {\mathbf{R}^i}^{\mathbf{T}}\mathbf{b}^{i}_{global}=
{\mathbf{R}^i}^{\mathbf{T}}(\mathbf{p}^{child(i)} - \mathbf{p}^{parent(i)})
\end{equation}
For convenience, this local bone vector is further converted into spherical coordinates $\mathbf{b}^{i}_{local} = (r_{i}, \theta_{i}, \phi_{i})$. The posture of the skeleton is described by the collection of bone orientations $\{(\theta_i, \phi_i)\}_{i=1}^w$ while the skeleton size is encoded into $\{r_{i}\}_{i=1}^w$.

\begin{figure*}[ht]
	\begin{center}
		\includegraphics[width=1\linewidth, trim=2cm 2cm 2cm 2cm]{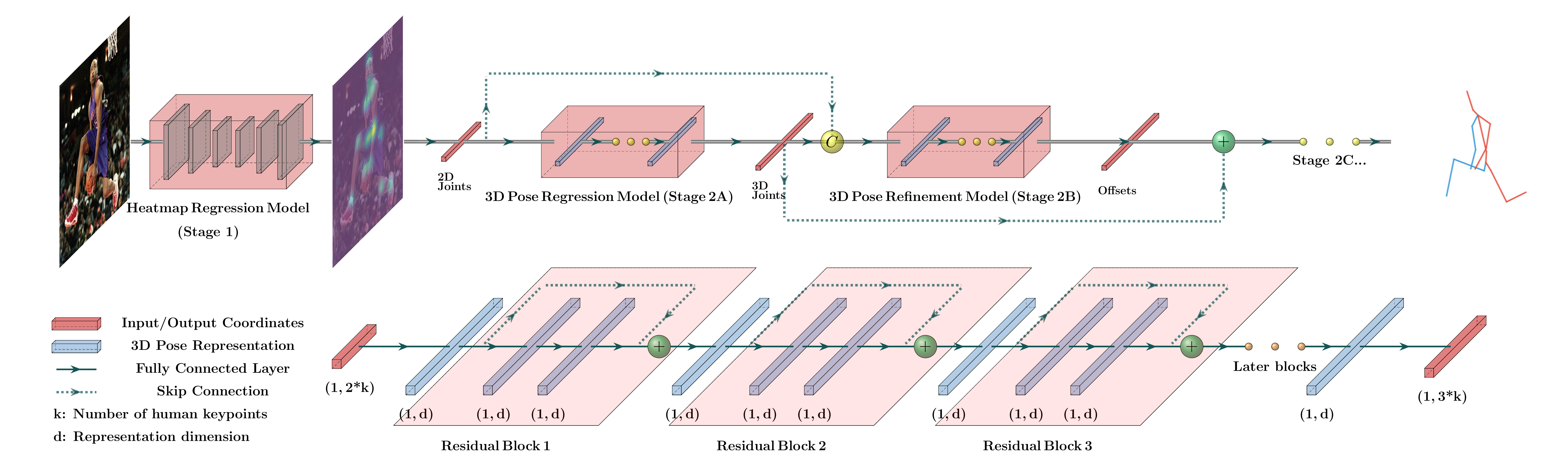}
	\end{center}
	\caption{Our cascaded 3D pose estimation architecture. Top: our model is a two-stage model where the first stage is a 2D landmark detector and the second stage is a cascaded 3D coordinate regression model. Bottom: each learner in the cascade is a feed-forward neural network whose capacity can be adjusted by the number of residual blocks. To fit an evolved dataset with plenty 2D-3D pairs, we use 8 layers (3 blocks) for each deep learner and have 24 layers in total with a cascade of 3 deep learners.}
	\label{model architecture}
\end{figure*}
%-------------------------------------------------------------------------

\subsection{Synthesizing New 2D-3D Pairs}
We first synthesize new 3D skeletons $\mathcal{D}_{new} = \{\mathbf{p}_j\}_{j=1}^{M}$ with an initial training dataset $\mathcal{D}_{old} = \{\mathbf{p}_i\}_{i=1}^{N}$ and project 3D skeletons to 2D given camera intrinsics $\mathbf{K}$ to form 2D-3D pairs $\{(\phi(\mathbf{x}_j), \mathbf{p}_j)\}_{i=1}^{M}$ where $\phi(\mathbf{x}_j) = \mathbf{K}\mathbf{p}_j$.

When adopting the hierarchical representation, a dataset of articulated 3D objects is a population of tree-structured data in nature. Evolutionary operators~\cite{holland1992adaptation} have \emph{constructive} property~\cite{spears1993crossover} that can be used to synthesize new data~\cite{correia2019evolutionary} given an initial population. The design of operators is problem-dependent and our operators are detailed as follows.

\noindent\textbf{Crossover Operator} Given two parent 3D skeletons represented by two kinematic trees, crossover is defined as a random exchange of sub-trees. This definition is inspired by the observation that an unseen 3D pose might be obtained by assembling limbs from known poses. Formally, we denote the set of bone vectors for parent $A$ and $B$ as $S_{A} = \{\mathbf{b}_A^1, \mathbf{b}_A^2, \dots, \mathbf{b}_A^w\}$ and $S_{B} = \{\mathbf{b}_B^1, \mathbf{b}_B^2, \dots, \mathbf{b}_B^w\}$. A joint indexed by $q$ is selected at random and the bones rooted at it are located for the two parents. These bones form the chosen sub-tree set $S_{chosen}$
\begin{equation}
\{\mathbf{b}^{j}:parent(j)=q \lor \mbox{\textit{IsOff}}(parent(j), q)\}
\end{equation}
where  $\mbox{\textit{IsOff}}(parent(j), q)$ is $\mathit{True}$ if joint $parent(j)$ is an offspring of joint $q$ in the kinematic tree. The parent bones are split into the chosen and the remaining ones as
$S_{X} = S_{chosen}^X \cup S_{rem}^X$where $S_{rem}^X = S_{X} - S_{chosen}^X$ and $X$ is $A$ or $B$. Now the crossover operator gives two sets of children bones as 
\begin{equation}
S_{C} = S_{chosen}^A \cup S_{rem}^B\quad \textrm{and} \quad S_{D} = S_{chosen}^B \cup S_{rem}^A
\end{equation}
These two new sets are converted into two new 3D skeletons. The example in Fig.~\ref{fig:data generation} shows the exchange of the right arms when the right shoulder joint is selected.
\begin{algorithm}[t]
	\footnotesize
	\caption{Data evolution} 
	\hspace*{0.02in} {\bf Input:} \\
	\hspace*{0.02in} Initial set of 3D skeletons $D_{old} = \{\mathbf{p}_i\}_{i=1}^{N}$, noise level $\sigma$, number of generations $G$\\
	\hspace*{0.02in} {\bf Output:} 
	Augmented set of skeletons $D_{new} = \{\mathbf{p}_i\}_{i=1}^{M}$
	\begin{algorithmic}[1]
		\State $D_{new} = D_{old}$
		\For{i=1:$G$} 
		\State Parents = Sample($D_{new}$)
		\State $\mbox{Children} = \mbox{NaturalSelection}(\mbox{Mutation(Crossover(Parents))})$
		\State $D_{new} = D_{new} \cup \mbox{Children}$
		\EndFor
		\State \Return $D_{new}$
	\end{algorithmic}
	\label{evolution}
	
\end{algorithm}

\noindent\textbf{Mutation Operator} As the motion of human limbs is usually continuous, a perturbation of one limb of an old 3D skeleton may result in a valid new 3D pose. To implement this perturbation, our mutation operator modifies the local orientation of one bone vector to get a new pose. One bone vector $\mathbf{b}_{i} = (r_{i}, \theta_{i}, \phi_{i})$ for an input 3D pose is selected at random and its orientation is mutated by adding noise (Gaussian in this study):
\begin{equation}
\theta_{i}' = \theta_{i} + g_{\theta}, \phi_{i}' = \phi_{i} + g_{\phi}
\end{equation}
where $g \sim N(0, \sigma)$ and $\sigma$ is a pre-defined noise level. One example of mutating the left leg is shown in Fig.~\ref{fig:data generation}. We also mutate the global orientation and bone length of the 3D skeletons to reduce the data bias of viewpoints and subject sizes, which is detailed in our supplementary material.

\noindent\textbf{Natural Selection}  
We use a fitness function to evaluate the goodness of synthesized data for selection as $v(\mathbf{p})$ which indicates the validity of the new pose. $v(\mathbf{p})$ can be any function that describes how anatomically valid a skeleton is, and we implement it by utilizing the binary function in~\cite{akhter2015pose}. We specify $ v(\mathbf{p})=-\infty$ if $\mathbf{p}$ is not valid to rule out all invalid poses.

\noindent\textbf{Evolution Process} The above operators are applied to $D_{old}$ to obtain a new generation $D_{new}$ by synthesizing new poses and merge with the old poses. This evolution process repeats $G$ generations and is depicted in Algorithm~\ref{evolution}. Finally, $D_{new}$ are projected to 2D key-points to obtain paired 2D-3D supervision. 

\section{Model Architecture}
We propose a two-stage model as shown in Fig.~\ref{model architecture}. We name it TAG-Net, as the model's focus \textbf{t}ransits from \textbf{a}ppearance to \textbf{g}eometry. This model can be represented as a function

\begin{equation}
\hat{\mathbf{p}} = \mathit{TAG}(\mathbf{x}) = \mathcal{G}(\mathcal{A}(\mathbf{x}))
\end{equation}
Given an input RGB image $\mathbf{x}$, $\mathcal{A}(\mathbf{x})$ (the appearance stage) regresses $k=17$ high-resolution probability heat-maps $\mathbf{H}_{i=1}^{k}$ for $k$ 2D human key-points and map them into 2D coordinates $\mathbf{c} = (x_i, y_i)^{k}_{i=1}$. $\mathcal{G}(\mathbf{c})$ (the geometry stage) infers 3D key-point coordinates\footnote{Relative to the root joint.} $\mathbf{p} = (x_i, y_i, z_i)^{k}_{i=1}$ in the camera coordinate system from input 2D coordinates. Key designs are detailed as follows.

\subsection{High-resolution Heatmap Regression}
Synthesized 2D key-points are projected from 3D points and can be thought as perfect detections while real detections produced by heat-map regression models are more noisy. We hope this noise can be small since we need to merge these two types of data as described in Section~\ref{dataevo}. To detect 2D key-points as accurate as possible, we decide to obtain feature maps with high spatial resolution and use HR-Net~\cite{sun2019deep} as our backbone for feature extraction. While the original model predicts heat-maps of size 96 by 72, we append a pixel shuffle super-resolution layer \cite{shi2016real} to the end and regress heat-maps of size 384 by 288. The original model~\cite{sun2019deep} uses hard arg-max to predict 2D coordinates, which results in rounding errors in our experiments. Instead, we use soft arg-max \cite{nibali2018numerical, sun_2018_eccv} to obtain 2D coordinates. The average 2D key-point localization errors for H36M testing images are shown in Tab.~\ref{tab:acc_2d}. Our design choice improves the previous best model and achieves the \emph{highest} key-point localization accuracy on H36M to date. The extensions add negligible amount of parameters and computation.

\begin{table}[h]
	\footnotesize
	\centering
	\newcommand{\TableEntry}[2]{$\text{#1}_{\downarrow \text{#2}\%}$}
	\begin{tabular}{|l|c|c|c|c|}
		\hline
		Backbone & Extension& \#Params & FLOPs & Error$\downarrow$ \\
		\hline
		\rowcolor{grayDark}
		CPN~\cite{chen2018cascaded} &- &  - &13.9G  &5.40 \\
		\rowcolor{grayLight}
		HRN~\cite{sun2019deep}  &-&   63.6M&32.9G  & \TableEntry{4.98}{7.8}\\
		\rowcolor{grayDark}
		HRN& + U&   63.6M&32.9G  &\TableEntry{4.64}{14.1}\\	
		\rowcolor{grayLight}
		HRN& + U + S&   63.6M&32.9G  &\TableEntry{4.36}{19.2}\\			
		\hline
	\end{tabular}
	\caption{Average 2D key-point localization errors for H36M testing set in terms of pixels. U:~Heat-map up-sampling. S:~use soft-argmax. Error reduction compared with the previous best model~\cite{chen2018cascaded} used in~\cite{Pavllo_2019_CVPR} follows the $\downarrow$ signs.}
	\label{tab:acc_2d}
\end{table} 

\begin{figure*}[h]
	\centering
	\includegraphics[scale=0.35, trim=1cm 0cm 1cm 0cm]{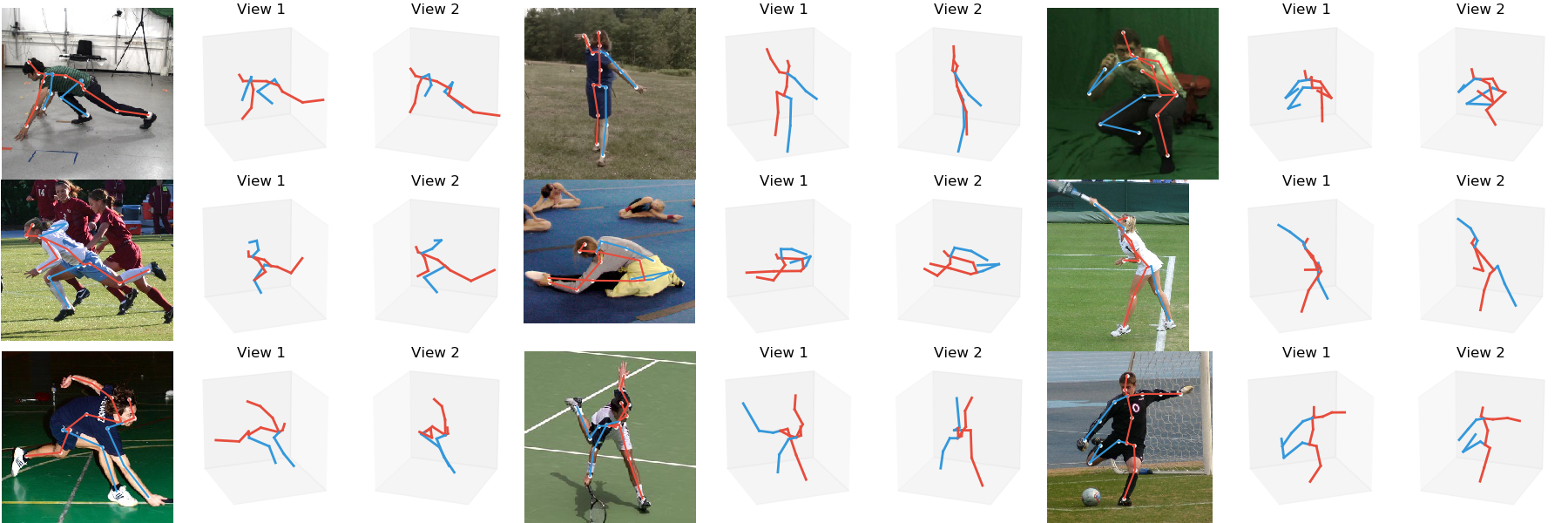}
	\caption{Cross-dataset inferences of $\mathcal{G}(\mathbf{c})$ on MPI-INF-3DHP (first row) and LSP (next two rows).}
	\label{qualitative}
\end{figure*}

\subsection{Cascaded Deep 3D Coordinate Regression}
Since the mapping from 2D coordinates to 3D joints can be highly-nonlinear and difficult to learn, we propose a cascaded 3D coordinate regression model as 
\begin{equation}
\hat{\mathbf{p}} = \mathcal{G}(\mathbf{c}) = \sum_{t=1}^{T}\mathcal{D}_{t}(\mathbf{i}_{t}, \mathbf{\Theta}_{t})
\end{equation}
where $\mathcal{D}_{t}$ is the $t$th \textit{deep} learner in the cascade parametrized by $\mathbf{\Theta}_{t}$ and takes input $\mathbf{i}_{t}$. As shown in the top of Fig.~\ref{model architecture}, the first learner $\mathcal{D}_{1}$ in the cascade directly predicts 3D coordinates while the later ones predict the 3D refinement $\delta \mathbf{p} = (\delta x_i, \delta y_i, \delta z_i)^{k}_{i=1}$. While cascaded coordinate regression has been adopted for 2D key-points localization \cite{cao2014face, Ren_2014_CVPR}, hand-crafted image feature and classical weak learners such as linear regressors were used. In contrast, our geometric model $\mathcal{G}(\mathbf{c})$ only uses coordinates as input and each learner is a fully-connected (FC) DNN with residual connections \cite{he2016deep}. 

The bottom of Fig.~\ref{model architecture} shows the detail for each deep learner. One deep learner first maps the input 2D coordinates into a representation vector of dimension $d=1024$, after which $R=3$ residual blocks are used. Finally the representation is mapped into 3D coordinates. After each FC layer we add batch normalization \cite{ioffe2015batch} and dropout \cite{srivastava2014dropout} with dropout rate 0.5. The capacity of each deep learner can be controlled by $R$. This cascaded model is trained sequentially by gradient descent and the training algorithm is included in our supplementary material. Despite the number of parameters increase linearly with the cascade length, we found that the cascaded model is robust to over-fitting for this 3D coordinate prediction problem, which is also shared by the 2D counterparts~\cite{cao2014face, Ren_2014_CVPR}.

\subsection{Implementation Details}
The camera intrinsics provided by H36M are used to project 3D skeletons. We train $\mathcal{A}(\mathbf{x})$ and $\mathcal{G}(\mathbf{c})$ sequentially. The input size is 384 by 288 and our output heat-map has the same high resolution. The back-bone of $\mathcal{A}(\mathbf{x})$ is pre-trained on COCO \cite{lin2014microsoft} and we fine-tune it on H36M with Adam optimizer using a batch size of 24. The training is performed on two NVIDIA Titan Xp GPUs and takes 8 hours for 18k iterations. We first train with learning rate 0.001 for 3k iterations, after which we multiply it by 0.1 after every 3k iterations. To train $\mathcal{G}(\mathbf{c})$, we train each deep learner in the cascade using Adam optimizer with learning rate 0.001 for 200 epochs.

\section{Experiments}
To validate our data evolution framework, we evolve from the training data provided in H36M and investigate how data augmentation may affect the generalization ability of 2D-to-3D networks. We conduct both intra- and cross-dataset evaluation. Intra-dataset evaluation is performed on H36M and demonstrates the model performance in an indoor environment similar to the training data. Cross-dataset evaluation is conducted on datasets not seen during training to simulate a larger domain shift. Considering the availability of MC data may vary in different application scenarios, We vary the size of initial population starting from scarce training data. These experiments help comparison with other weakly/semi-supervised methods that only use very few 3D annotation but do not consider data augmentation. Finally we present an ablation study to analyze the influences of architecture design and choice of hyper-parameters.
\subsection{Datasets and Evaluation Metrics}
\textbf{Human 3.6M} (H36M) is the largest 3D human pose estimation benchmark with accurate 3D labels. We denote a collection of data by appending subject ID to S, e.g., S15 denotes data from subject 1 and 5. Previous works fix the training data while our method uses it as our initial population and evolves from it. We evaluate model performance with \textit{Mean Per Joint Position Error} (MPJPE) measured in millimeters. Two standard evaluation protocols are adopted. \textit{Protocol 1} (P1) directly computes MPJPE while \textit{Protocol 2} (P2) aligns the ground-truth 3D poses with the predictions with a rigid transformation before calculating it. Protocol P1$^*$ uses ground truth 2D key-points as inputs and removes the influence of the first stage model. 

\textbf{MPI-INF-3DHP} (3DHP) is a benchmark that we use to evaluate the generalization power of 2D-to-3D networks in unseen environments. We do not use its training data and conduct cross-dataset inference by feeding the provided key-points to $\mathcal{G}(\mathbf{c})$. Apart from MPJPE, \textit{Percentage of Correct Keypoints} (PCK) measures correctness of 3D joint predictions under a specified threshold, while \textit{Area Under the Curve} (AUC) is computed for a range of PCK thresholds. 

\textbf{\underline{U}nconstrained \underline{3D} \underline{P}oses in the \underline{W}ild} (U3DPW) We collect by ourselves a new small dataset consisting of 300 challenging in-the-wild images with rare human poses, where 150 of them are selected from Leeds Sports Pose dataset~\cite{Johnson10}. The annotation process is detailed in our supplementary material. 
This dataset is used for qualitatively validating model generalization for unseen rare poses.

\subsection{Comparison with state-of-the-art methods}

\noindent \textbf{Comparison with weakly-supervised methods}. Here we compare with \emph{weakly/semi-supervised methods}, which only use a small number of training data to simulate scarce data scenario. To be consistent with others,  we utilize S1 as our initial population. While others fix S1 as the training dataset, we evolve from it to obtain an augmented training set. The comparison of model performance is shown in Tab.~\ref{tab:h36m_weakly}, where our model significantly out-performs others and demonstrates effective use of the limited training data. While other methods \cite{Rhodin_2018_CVPR, kocabas2019self} use multi-view consistency as extra supervision, we achieve comparable performance with only a single view by synthesizing useful supervision. Fig.~\ref{wsr} validates our method when the training data is extremely scarce, where we start with a small fraction of S1 and increase the data size by 2.5 times by evolution. Note that the model performs consistently better after dataset evolution. Compared to the temporal convolution model proposed in \cite{Pavllo_2019_CVPR}, we do not utilize any temporal information and achieve comparable performance. This indicates our approach can make better use of extremely limited data.

\begin{table}[h]
	\footnotesize
	\centering
	\begin{tabular}{|c|c|c|c|}
		\hline
		\multirow{2}{*}{Method (Reference)} & \multicolumn{3}{c|}{Average MPJPE$\downarrow$} \\
		\cline{2-4}
		 & P1 & P1*  & P2 \\
		\hline 
		\multicolumn{4}{|c|}{Use Multi-view Images}\\
		\hline 
		\rowcolor{grayDark}
		Rhodin \etal~(CVPR'18) \cite{Rhodin_2018_CVPR}&   -& - &64.6  \\		
		\rowcolor{grayLight}
		Kocabas \etal~(CVPR'19) \cite{kocabas2019self} &  65.3& - &57.2  \\
		\hline
		\multicolumn{4}{|c|}{Use Temporal Information from Videos}\\
		\hline 		
		\rowcolor{grayDark}
		Pavllo \etal~(CVPR'19) \cite{Pavllo_2019_CVPR}&64.7 & - & -  \\		
		\hline 
		\multicolumn{4}{|c|}{Use a Single RGB Image}\\
		\hline 		
		\rowcolor{grayDark}
		Li \etal~(ICCV'19) \cite{Li_2019_ICCV}&88.8 & - & 66.5  \\	
		\hline
		Ours~(CVPR' 20) &\textbf{60.8} &\textbf{50.5}  &\textbf{46.2} \\				
		\hline
	\end{tabular}
	
	\caption{Comparison with SOTA weakly-supervised methods. Average MPJPE over all 15 actions for H36M under two protocols (P1 and P2) is reported. P1* refers to protocol 1 evaluated with ground truth 2d key-points. Best performance is marked with bold font. Error for each action can be found in Tab.~\ref{table:extend_h36m_weakly}. }
	\label{tab:h36m_weakly}
\end{table} 

\noindent \textbf{Comparison with fully-supervised methods}. Here we compare with \emph{fully-supervised methods} that uses the whole training split of H36M. We use S15678 as our initial population and Tab.~\ref{tab:h36m_fully} shows the performance comparison. Under this setting, our model also achieves competitive performance compared with other SOTA methods, indicating that our approach is not limited to scarce data scenario.

\begin{table}[h]
	\footnotesize
	\centering
	\begin{tabular}{|c|c|c|c|}
		\hline
		\multirow{2}{*}{Method (Reference)} & \multicolumn{3}{c|}{Average MPJPE$\downarrow$} \\
		\cline{2-4}
		& P1 & P1*  & P2 \\
		\hline
		\rowcolor{grayDark}
		Martinez~\etal~(ICCV'17)~\cite{martinez2017simple} &62.9 & 45.5 & 47.7  \\
		\rowcolor{grayLight}
		Yang~\etal~(CVPR'18)~\cite{yang20183d}   & 58.6& - & \textbf{37.7}  \\			
		\rowcolor{grayDark}
		Zhao~\etal~(CVPR'19)~\cite{Zhao_2019_CVPR} &57.6 & 43.8 & -  \\		
		\rowcolor{grayLight}
		Sharma~\etal~(ICCV'19)~\cite{Sharma_2019_ICCV} &  58.0 & - & 40.9 \\
		\rowcolor{grayDark}
		Moon~\etal~(ICCV'19)~\cite{Moon_2019_ICCV_3DMPPE}  &  54.4 & 35.2 &-  \\
		\hline
		Ours~(CVPR' 20)  &\textbf{50.9} & \textbf{34.5}  &38.0  \\	
		\hline
	\end{tabular}
	
	\caption{Comparison with SOTA methods under fully-supervised setting. Same P1, P1* and P2 as in Tab.~\ref{tab:h36m_weakly}. Error for each action can be found in Tab.~\ref{table:extend_h36m_fully}.}
	\label{tab:h36m_fully}
\end{table} 

%-------------------------------------------------------------------------
\subsection{Cross-dataset Generalization}
\begin{figure}
	\centering
	\begin{minipage}{.5\linewidth}
		\centering
		\subfloat{\label{a}\includegraphics[scale=.2, trim=0.5cm 1cm 0.5cm 1cm]{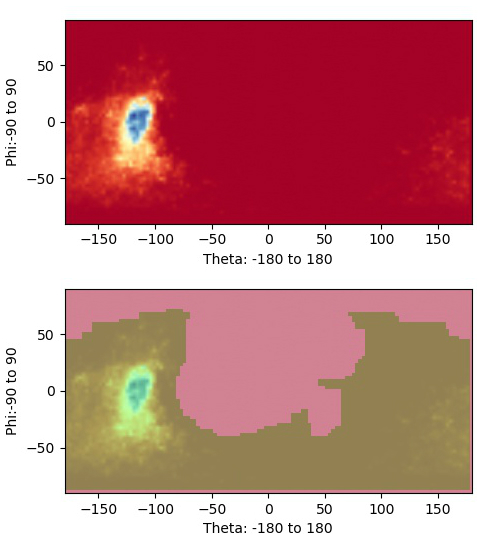}}
	\end{minipage}%\par\medskip	
	\begin{minipage}{.5\linewidth}
		\centering
		\subfloat{\label{b}\includegraphics[scale=.2, trim=0.5cm 1cm 0.5cm 1cm]{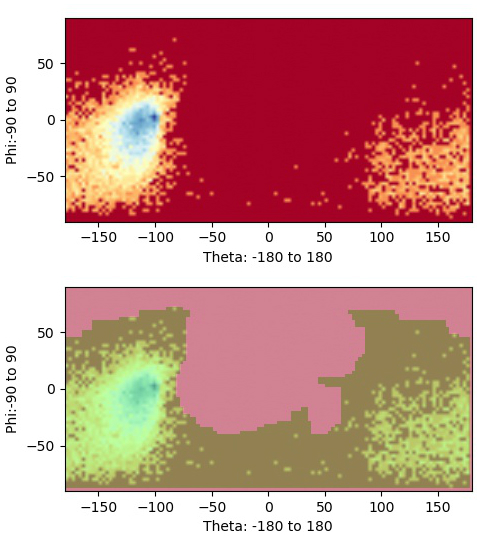}}
	\end{minipage}%\par\medskip	
	\caption{Dataset distribution for the bone vector connecting right shoulder to right elbow. Top: distribution before (left) and after (right) dataset augmentation. Bottom: distribution overlaid with valid regions (brown) taken from \cite{akhter2015pose}.}
	\label{distribution}
\end{figure}

\begin{figure*}[h]
	\centering
	\includegraphics[scale=0.35, trim=1cm 0cm 1cm 0cm]{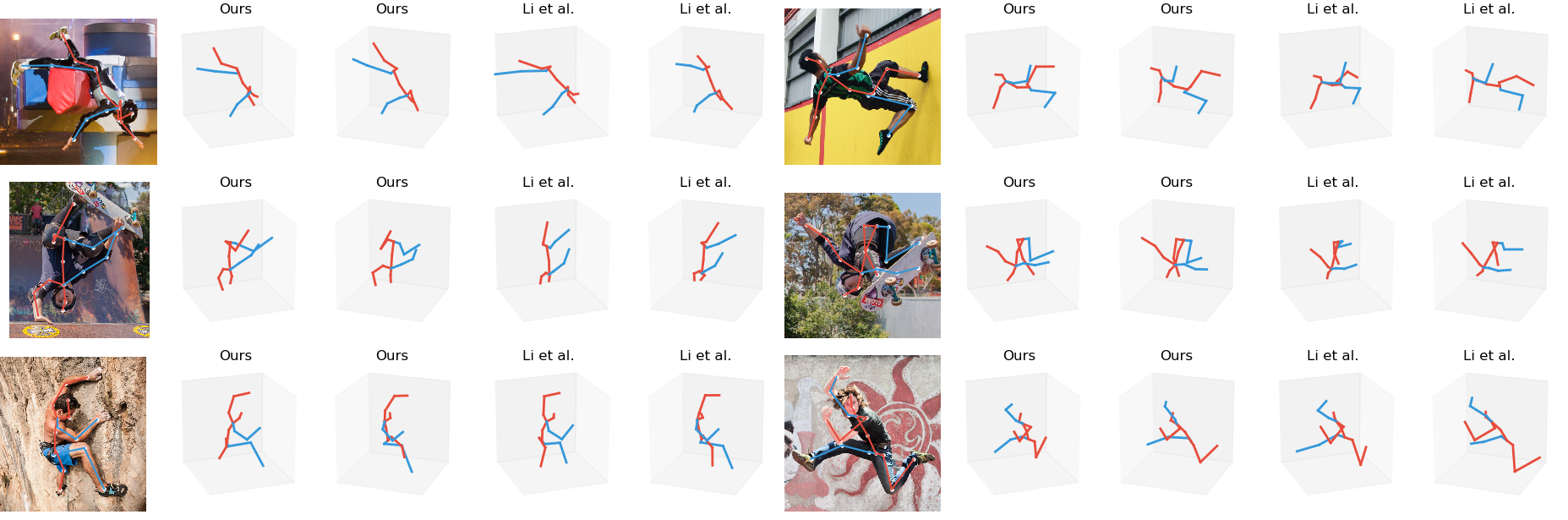}
	\caption{Cross-dataset inference results on U3DPW comparing with~\cite{Li_2019_CVPR}. Video is included in our supplementary material.}
	\label{qualitative_comp}
\end{figure*}

To validate the generalization ability of our 2D-to-3D network in unknown environment, Tab.~\ref{tab:3dhp} compares with other methods on 3DHP. In this experiment we evolve from S15678 in H36M to obtain an augmented dataset consisting of 8 million 2D-3D pairs. Without utilizing any training data of 3DHP, $\mathcal{G}(\mathbf{c})$ achieves competitive performance in this benchmark. We obtain clear improvements comparing with \cite{Li_2019_CVPR}, which also uses S15678 as the training data but fix it without data augmentation. The results indicate that our data augmentation approach improves model generalization effectively despite we start with the same biased training dataset. As shown in Fig.~\ref{distribution}, the distribution of the augmented dataset indicates less dataset bias. Qualitative results on 3DHP and LSP are shown in Fig.~\ref{qualitative}. Note that these unconstrained poses are not well-represented in the original training dataset yet our model still gives good inference results. Qualitative comparison with \cite{Li_2019_CVPR} on some difficult poses in U3DPW is shown in Fig.~\ref{qualitative_comp} and our $\mathcal{G}(\mathbf{c})$ generalizes better for these rare human poses.

\begin{table}[h]
	\footnotesize
	\centering
	\begin{tabular}{|l|c|c|c|c|}
		\hline
		Method & CE & PCK$\uparrow$ & AUC$\uparrow$ & MPJPE$\downarrow$ \\
		\hline
		\rowcolor{grayDark}
		Mehta et al.  \cite{mehta2017monocular} & & 76.5 & 40.8 & 117.6 \\
		\rowcolor{grayLight}
		VNect  \cite{mehta2017vnect} &  & 76.6 & 40.4 & 124.7 \\
		\rowcolor{grayDark}
		LCR-Net  \cite{rogez2017lcr} & & 59.6 & 27.6 & 158.4 \\		
		\rowcolor{grayLight}
		Zhou et al.  \cite{zhou2017towards} & & 69.2 & 32.5 & 137.1 \\
		\rowcolor{grayDark}
		Multi Person  \cite{mehta2018single} & & 75.2 & 37.8 & 122.2 \\
		\rowcolor{grayLight}
		OriNet  \cite{luo2018orinet} & & 81.8 & 45.2 & \textbf{89.4} \\
		\hline
		\rowcolor{grayDark}
		Li et al.  \cite{Li_2019_CVPR}  &\checkmark  & 67.9  & - & -\\
		\rowcolor{grayLight}		
		Kanazawa  \cite{kanazawa2018end} & \checkmark & 77.1 &	40.7 & 113.2 \\
		\rowcolor{grayDark}
		Yang et al.  \cite{yang20183d} & \checkmark & 69.0 & 32.0 & - \\
		\hline			
		Ours  & \checkmark &   \textbf{81.2}   & \textbf{46.1}  & \textbf{99.7} \\
		\hline
	\end{tabular}
	\caption{Testing results for the MPI-INF-3DHP dataset. A higher value is better for PCK and AUC while a lower value is better for MPJPE. MPJPE is evaluated without rigid transformation. CE denotes cross-dataset evaluation and the training data in MPI-INF-3DHP is not used.}
	\label{tab:3dhp}
\end{table}

\subsection{Ablation Study}
\noindent \textbf{Effect of data evolution}. Our ablation study is conducted on H36M and summarized in Tab.~\ref{tab:ablation}. The baseline (B) uses $T$=1. Note that adding cascade (B+C) and dataset evolution (B+C+E) consistently out-performs the baseline. Discussion on the evolution operators is included in our supplementary material.

\begin{figure}[h]
	\centering
	\includegraphics[scale=0.25, trim=1cm 1.5cm 1cm 1cm]{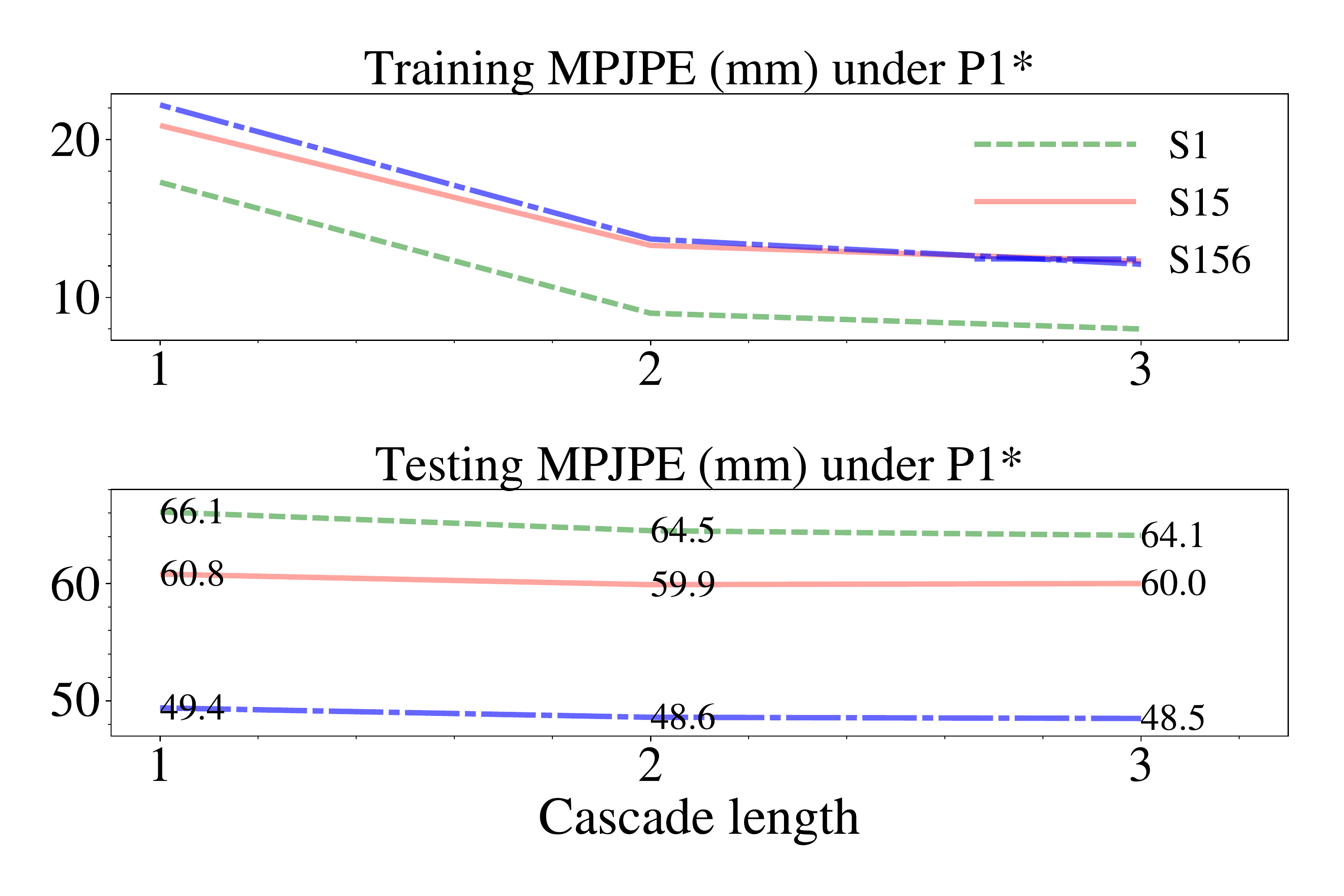}
	\caption{Training and testing errors with varying number of cascade length and training data. S156 indicates using 3D annotations from subjects 1, 5 and 6. The cascade effectively reduces training error and is robust to over-fitting.}
	\label{cascade}
\end{figure}

\noindent \textbf{Effect of cascade length T}. Here we train our model on various subsets of H36M and plot MPJPE over cascade length as shown in Fig.~\ref{cascade}. Here $R$ is fixed as 2. Note that the training error increases as the training set becomes more complex and the testing errors decreases accordingly. The gap between these two errors indicate insufficient training data. Note that with increasing number of deep learners, the training error is effectively reduced but the model does not overfit. This property is brought by the ensemble effect of multiple deep learners.  

\begin{figure}[h]
	\centering
	\includegraphics[scale=0.3, trim=2cm 1.5cm 2cm 2cm]{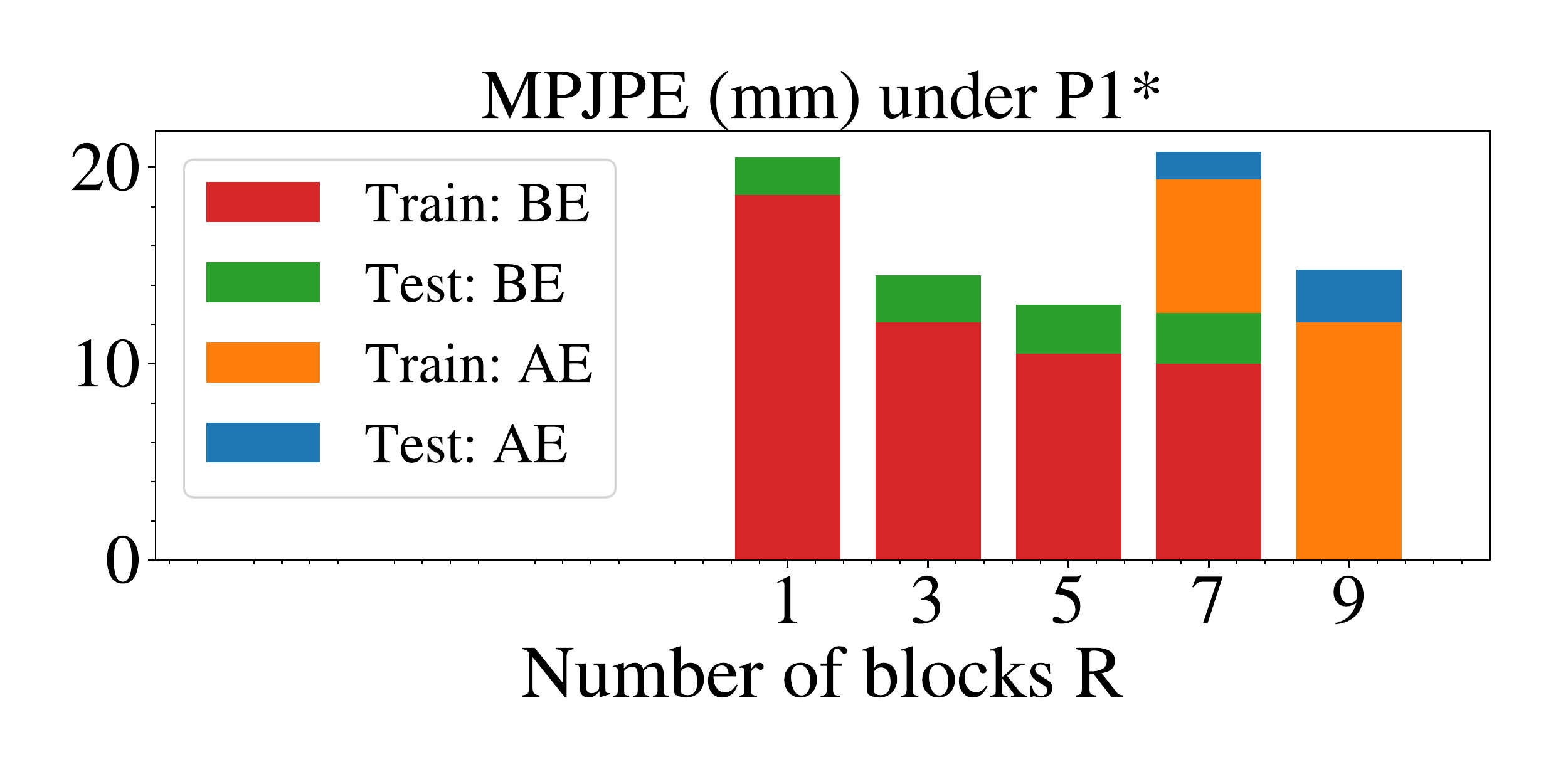}
	\caption{MPJPE (P1*) before (BE) and after evolution (AE) with varying number of blocks $R$. Evolved training data can afford a deeper network. Best viewed in color.}
	\label{block}
\end{figure}

\noindent \textbf{Effect of block number R}. Here we fix $T$=1, $d$=512 and vary $R$. S15678 in H36M and its evolved version are used. The datasets before (BE) and after evolution (AE) are randomly split into training and testing subsets for simplicity. The training and testing MPJPEs are shown in Fig.~\ref{block}. Note that the training error is larger after evolution with the same $R$=7. This means our approach increases the variation of training data, which can afford a deeper architecture with larger $R$ (e.g. $R$=9). 

\begin{table}[ht]
	\footnotesize
	\centering
	\newcommand{\TableEntry}[2]{$\text{#1}_{\downarrow \text{#2}\%}$}
	\begin{tabular}{|l|c|c|c|}
		\hline
		\multirow{2}{*}{Method} &  \multirow{2}{*}{Training Data} & \multicolumn{2}{c|}{MPJPE}\\ \cline{3-4}
		 &   & P1 & P1*\\
		\hline
		\multicolumn{4}{|c|}{Problem Setting A: Weakly-supervised Learning}\\
		\hline 
		\rowcolor{grayDark}
		B & S1 & 71.5   &66.2\\
		\rowcolor{grayLight}
		B+C  & S1 & \TableEntry{70.1}{2.0}  &\TableEntry{64.5}{2.6}\\
		\rowcolor{grayDark}
		B+C+E& Evolve(S1) &\TableEntry{60.8}{15.0}  &\TableEntry{50.5}{21.7}\\	
		\hline
		\multicolumn{4}{|c|}{Problem Setting B: Fully-supervised Learning}\\
		\hline
		\rowcolor{grayDark}
		B  &  S15678 & 54.3 &44.5\\
		\rowcolor{grayLight}
		B+C  & S15678 &  \TableEntry{52.1}{4.0} &\TableEntry{42.9}{3.6}\\
		\rowcolor{grayDark}
		B+C+E& Evolve(S15678) & \TableEntry{50.9}{6.2}  &\TableEntry{34.5}{22.4}\\				
		\hline
	\end{tabular}
	\caption{Ablation study on H36M. B:~baseline. C:~add cascade. E:~add data evolution. Evolve() represents the data augmentation operation. Same P1 and P1* as in Table \ref{tab:h36m_weakly}. Error reduction compared with the baseline follows the $\downarrow$ signs.}
	\label{tab:ablation}
\end{table}

\section{Conclusion} 
This paper presents an evolutionary framework to enrich the 3D pose distribution of an initial biased training set. This approach leads to better intra-dataset and cross-dataset generalization of 2D-to-3D network especially when available 3D annotation is scarce. A novel cascaded 3D human pose estimation model is trained achieving state-of-the-art performance for single-frame 3D human pose estimation. There are many fruitful directions remaining to be explored. Extension to temporal domain, multi-view setting and multi-person scenarios are three examples. In addition, instead of being fixed, the operators can also evolve during the data generation process.  
 
\noindent \textbf{Acknowledgments} We gratefully acknowledge the support
of NVIDIA Corporation with the donation of one Titan Xp GPU used for this research. This research is also supported in part by Tencent and the Research Grant Council of the Hong Kong SAR under grant no. 1620818.

\clearpage
% supplementary material
\begin{center}
	\large{\textbf{Supplementary Material}}
\end{center}

%%%%%%%%% BODY TEXT
This supplementary material includes implementation details and extended experimental analysis that are not included in the main text due to space limit. The detailed MPJPE under different settings are shown Tab.~\ref{table:extend_h36m_fully} and Tab.~\ref{table:extend_h36m_weakly}. Other contents are organized in separate sections as follows:
\begin{itemize}
	\item \hyperref[sec1]{\textbf{Section \ref{sec1}}} includes the implementation details of the \textbf{hierarchal human representation}.
	\item \hyperref[sec2]{\textbf{Section \ref{sec2}}} elaborates the \textbf{model training}, which includes the training algorithm of the cascaded model and describes details of data pre-processing.
	\item \hyperref[sec3]{\textbf{Section \ref{sec3}}} gives \textbf{ablation study} on data generation and the evolutionary operators.
	\item \hyperref[sec4]{\textbf{Section \ref{sec4}}} describes the new \textbf{dataset U3DPW} and its collection process.
\end{itemize}

\begin{table*}[t]
	\centering
	% \small
	\resizebox{\textwidth}{!}{
		\begin{tabular}{@{}l|ccccccccccccccc|c@{}}
			\toprule[1pt]
			\textbf{Protocol \#1} & Dir. & Disc & Eat & Greet & Phone & Photo & Pose & Purch. & Sit & SitD. & Smoke & Wait & WalkD. & Walk & WalkT. & Avg.\\
			\midrule[0.5pt]
			
			Martinez \emph{et al.} (ICCV'17)\cite{martinez2017simple} &51.8 &56.2 &58.1 &59.0 &69.5 &78.4 &55.2 &58.1 &74.0 &94.6 &62.3 &59.1 &65.1 &49.5 &  52.4 &62.9\\
			
			Fang \emph{et al.} (AAAI'18) \cite{fang2018learning} & 50.1& 54.3& 57.0& 57.1& 66.6& 73.3& 53.4& 55.7& 72.8& 88.6& 60.3& 57.7& 62.7& 47.5& 50.6& 60.4 \\
			Yang \emph{et al.} (CVPR'18) \cite{yang20183d} &51.5 &58.9 &50.4 &57.0 &62.1 &65.4 &49.8 &52.7 &69.2 &85.2 &57.4 &58.4 &43.6 &60.1 &47.7 &58.6 \\

			Pavlakos \emph{et al.} (CVPR'18) \cite{pavlakos2018ordinal}&48.5 &54.4 &54.4 &52.0 &59.4 &65.3 &49.9 &52.9 &65.8 &71.1 &56.6 &52.9 &60.9 &44.7 &47.8 &56.2 \\
			
			Lee \emph{et al.} (ECCV'18) \cite{lee2018propagating} &\textbf{40.2} &49.2 &47.8 &52.6 &\textbf{50.1} &75.0 &50.2 &43.0 &55.8 &73.9 &54.1 &55.6 &58.2 &43.3 &\textbf{43.3} &52.8 \\

			Zhao \emph{et al.} (CVPR'19) \cite{Zhao_2019_CVPR}&47.3 &60.7 &51.4 &60.5 &61.1 &49.9 &\textbf{47.3} &68.1 &86.2 &55.0 &67.8 &61.0 &\textbf{42.1} &60.6 &45.3 &57.6 \\
					
			Sharma \emph{et al.} (ICCV'19) \cite{sharma2019monocular} &48.6 &54.5 &54.2 &55.7 &62.6 &72.0 &50.5 &54.3 &70.0 &78.3 &58.1 &55.4 &61.4 &45.2 &49.7 &58.0\\
						
			Moon \emph{et al.} (ICCV'19) \cite{Moon_2019_ICCV_3DMPPE} &51.5 &56.8 &51.2 &52.2 &55.2 &\textbf{47.7} &50.9 &63.3 &69.9 &\textbf{54.2} &57.4 &50.4 &42.5 &57.5 &47.7 &54.4\\		
				
			Liu \emph{et al.} (ECCV'20) \cite{liu2020comprehensive}&46.3 &52.2 &\textbf{47.3} &50.7 &55.5 &67.1 &49.2 &46.0 &60.4 &71.1 &51.5 &50.1 &54.5 &\textbf{40.3} &43.7 &52.4 \\
			
			\midrule[0.5pt]
			Ours:~Evolve(S15678) &47.0 &\textbf{47.1} &49.3 &\textbf{50.5} &53.9 &58.5 &48.8 &\textbf{45.5} &\textbf{55.2} &68.6 &\textbf{50.8} &\textbf{47.5} &53.6 &42.3 &45.6 &\textbf{50.9} \\
			
			\toprule[1pt]
			\textbf{Protocol \#2} & Dir. & Disc & Eat & Greet & Phone & Photo & Pose & Purch. & Sit & SitD. & Smoke & Wait & WalkD. & Walk & WalkT. & Avg.\\
			\midrule[0.5pt]
			
			Martinez \emph{et al.} (ICCV'17) \cite{martinez2017simple} &39.5 &43.2 &46.4 &47.0 &51.0 &56.0 &41.4 &40.6 &56.5 &69.4 &49.2 &45.0 &49.5 &38.0 &43.1 &47.7\\
			
			Fang \emph{et al.} (AAAI'18) \cite{fang2018learning} &38.2 &41.7 &43.7 &44.9 &48.5 &55.3 &40.2 &38.2 &54.5 &64.4 &47.2 &44.3 &47.3 &36.7 &41.7 &45.7 \\
			
			Pavlakos \emph{et al.} (CVPR'18) \cite{pavlakos2018ordinal}&34.7 &39.8 &41.8 &\textbf{38.6} &42.5 &47.5 &38.0 &36.6 &50.7 &56.8 &42.6 &39.6 &43.9 &32.1 &36.5 &41.8 \\

			Yang \emph{et al.} (CVPR'18) \cite{yang20183d} &\textbf{26.9} &\textbf{30.9} &\textbf{36.3} &39.9 &43.9 &47.4 &\textbf{28.8} &\textbf{29.4} &\textbf{36.9} &58.4 &41.5 &\textbf{30.5} &\textbf{29.5} &42.5 &\textbf{32.2} &\textbf{37.7} \\

			Sharma \emph{et al.} (ICCV'19) \cite{sharma2019monocular} &35.3 &35.9 &45.8 &42.0 &40.9 &52.6 &36.9 &35.8 &43.5 &51.9 &44.3 &38.8 &45.5 &\textbf{29.4} &34.3 &40.9\\
												
			Cai \emph{et al.} (ICCV'19) \cite{cai2019exploiting}&35.7 &37.8 &36.9 &40.7 &39.6 &45.2 &37.4 &34.5 &46.9 &\textbf{50.1} &40.5 &36.1 &41.0 &29.6 &33.2 &39.0 \\
			
			Liu \emph{et al.} (ECCV'20) \cite{liu2020comprehensive}&35.9 &40.0 &38.0 &41.5 &42.5 &51.4 &37.8 &36.0 &48.6 &56.6 &41.8 &38.3 &42.7 &31.7 &36.2 &41.2 \\			
			\midrule[0.5pt]
			Ours:~Evolve(S15678) &34.5&34.9&37.6&39.6&\textbf{38.8}&\textbf{45.9}&34.8&33.0&40.8&51.6&\textbf{38.0}&35.7&40.2&30.2&34.8&38.0 \\
			\toprule[1pt]
		\end{tabular}
	}
	\caption
	{
		Quantitative comparisons with the state-of-the-art fully-supervised methods on Human3.6M under protocol \#1 and protocol \#2. Best performance is indicated by bold font.
	}
	\label{table:extend_h36m_fully}
\end{table*}

\begin{table*}[t]
	\centering
	% \small
	\resizebox{\textwidth}{!}{
		\begin{tabular}{@{}l|ccccccccccccccc|c@{}}
			\toprule[1pt]
			\textbf{Protocol \#1} & Dir. & Disc & Eat & Greet & Phone & Photo & Pose & Purch. & Sit & SitD. & Smoke & Wait & WalkD. & Walk & WalkT. & Avg.\\
			\midrule[0.5pt]
			
			Kocabas \emph{et al.} (CVPR'19) \cite{kocabas2019self} &- &- &- &- &- &- &- &- &- &- &- &- &- &- &- &65.3 \\
			
			Pavllo \emph{et al.} (CVPR'19) \cite{Pavllo_2019_CVPR} &- &- &- &- &- &- &- &- &- &- &- &- &- &- &- &64.7 \\
			
			Li \emph{et al.} (ICCV'19) \cite{Li_2019_ICCV}&70.4 &83.6 &76.6 &78.0 &85.4 &\textbf{106.1} &72.2 &103.0 &115.8 &165.0 &82.4 &74.3 &94.6 &60.1 &70.6 &88.8 \\
			
			\midrule[0.5pt]
			Ours:~Evolve(S1) &\textbf{52.8} &\textbf{56.6} &\textbf{54.0} &\textbf{57.5} &\textbf{62.8} &\textbf{72.0} &\textbf{55.0} &\textbf{61.3} &\textbf{65.8} &\textbf{80.7} &\textbf{59.0} &\textbf{56.7} &\textbf{69.7} &\textbf{51.6} &\textbf{57.2} &\textbf{60.8} \\
			
			\toprule[1pt]
			\textbf{Protocol \#2} & Dir. & Disc & Eat & Greet & Phone & Photo & Pose & Purch. & Sit & SitD. & Smoke & Wait & WalkD. & Walk & WalkT. & Avg.\\
			\midrule[0.5pt]
			
			Rhodin \emph{et al.} (CVPR'18) \cite{Rhodin_2018_CVPR} &- &- &- &- &- &- &- &- &- &- &- &- &- &- &- &64.6\\
			
			Kocabas \emph{et al.} (CVPR'19) \cite{kocabas2019self} &- &- &- &- &- &- &- &- &- &- &- &- &- &- &- &57.2 \\
			
			Li \emph{et al.} (ICCV'19) \cite{Li_2019_ICCV}&- &- &- &- &- &- &- &- &- &- &- &- &- &- &- &66.5 \\
			
			\midrule[0.5pt]
			Ours:~Evolve(S1) &\textbf{40.1} &\textbf{43.4} &\textbf{41.9} &\textbf{46.1} &\textbf{48.2} &\textbf{55.1} &\textbf{42.8} &\textbf{42.6} &\textbf{49.6} &\textbf{61.1} &\textbf{44.5} &\textbf{43.2} &\textbf{51.5} &\textbf{38.0} &\textbf{44.4} &\textbf{46.2}  \\
			\toprule[1pt]
		\end{tabular}
	}
	\caption
	{
		Quantitative comparisons with the state-of-the-art weakly/semi-supervised methods on Human3.6M under protocol \#1 and protocol \#2. Best performance is indicated by bold font.
	}
	\label{table:extend_h36m_weakly}
\end{table*}

\section{Hierarchical Human Model} \label{sec1}
\subsection{Choice of Local Coordinate System}
As mentioned at equation \ref{transformation} in section \ref{hmr}, each global bone vector is transformed into a local bone vector with respect to a coordinate system attached at a parent joint. In general, the choice of the coordinate system is arbitrary and our evolutionary operators do not depend on it. In implementation, we adopt the coordinate system proposed in \cite{akhter2015pose}, where the computation of basis vectors depends on the 3D joint position. For the bone vectors representing upper limbs (left shoulder to left elbow, right shoulder to right elbow, left hip to left knee, right hip to right knee), the basis vectors are computed based on several joints belonging to the human torso. For the bone vectors representing lower limbs (left elbow to left wrist, right elbow to right wrist, left knee to left ankle, right knee to right ankle), the basis vectors are computed from the parent bone vectors. 

Algorithm \ref{local} is adapted from \cite{akhter2015pose} and details the process of computing basis vectors and performing coordinate transformation. Bold name such as \textbf{rightShoulder} denotes the global position of the 3D skeleton joint. We define a bone vector's \emph{parent bone vector} as the bone vector whose end point is the starting point of it. An index mapping function $M(i)$ is introduced here that maps bone vector index $i$ to the index of its \emph{parent bone vector}. Consistent with the notations of the main text, we have $child(M(i)) = parent(i)$.  In implementation, we found that the joints used in \cite{akhter2015pose} have slightly different semantic meaning compared to the data provided by H36M. Thus we use the bone vector connecting the spine and thorax joints to approximate the \emph{backbone vector} used in in \cite{akhter2015pose} (\textbf{backBone} in algorithm \ref{local}). 

\begin{algorithm}[h]
	\footnotesize
	\caption{Computation of local bone vector} 
	\hspace*{0.02in} {\bf Input:} 
	\hspace*{0.02in} $i$th global bone vector $\mathbf{b}_{g} = \mathbf{b}^{i}_{global}$, constant 3D vector $\mathbf{a}$\\
	\hspace*{0.02in} {\bf Output:} 
	$i$th local bone vector $\mathbf{b}_{l} = \mathbf{b}^{i}_{local}$ with its local coordinate system $\mathbf{R}^{i}$
	\begin{algorithmic}[1]
		\State \textbf{backBone} = \textbf{Spine} - \textbf{Thorax}
		\If {$\mathbf{b}_{g}$ is upper limb}
		\State $\mathbf{v}_{1}$ = \textbf{rightShoulder} - \textbf{leftShoulder}
		\State $\mathbf{v}_{2}$ = \textbf{backBone}
		\ElsIf {$\mathbf{b}_{g}$ is lower limb}
		\State $\mathbf{v}_{1}$ = \textbf{rightHip} - \textbf{leftHip}
		\State $\mathbf{v}_{2}$ = \textbf{backBone}
		\Else 
		\State $\mathbf{v}_{1}$ = $\mathbf{b}^{M(i)}_{g}$
		\State $\mathbf{v}_{2}$ = $\mathbf{R}^{M(i)}\mathbf{a} \times \mathbf{v}_{1}$
		\EndIf
		\State $\mathbf{R}^{i} = GramSchmidt(\mathbf{v}_{1}, \mathbf{v}_{2}, \mathbf{v}_{1} \times \mathbf{v}_{2})$
		\State $\mathbf{b}_{l} = {\mathbf{R}^{i}}^{T} \mathbf{b}_{g} $
		\State \Return $\mathbf{b}_{l}$
	\end{algorithmic}
	\label{local}
	
\end{algorithm}

\subsection{Validity Function}
To implement $v(\mathbf{p})$, local bone vectors are first computed by Algorithm~\ref{local} and converted into spherical coordinates as $\mathbf{b}_{local}^{i} = (r_i, \theta_i, \phi_i)$. A pose $\mathbf{p}$ is then considered as the collection of bone orientations ${(\theta_i, \phi_i)}_{i=1}^{w}$. A function is provided by \cite{akhter2015pose} to decide the validity of each tuple $(\theta_i, \phi_i)$. We define a pose $\mathbf{p}$ to be anthropometrically valid if every tuple $(\theta_i, \phi_i)$ is valid:

\begin{align*}
v(\mathbf{p}) & = \begin{cases}
0, & \mbox{if } (\theta_i, \phi_i) \mbox{ is valid for i=1,2,...,w},\\
-\infty, & \mbox{else}.
\end{cases}
\end{align*}
The original code released by \cite{akhter2015pose} was implemented by MATLAB and we provide a Python implementation on our project website.

\section{Model Training} \label{sec2}
\subsection{Training Procedure of the Cascaded Model}
We train each deep learner in the cascade sequentially as depicted by algorithm \ref{alg:cascade}. The $TrainNetwork$ is a routine representing the training process of a single deep learner, which consists of forward pass, backward pass and network parameter update using Adam optimizer. Starting from the second deep learner, the inputs can also be concatenated with the current estimates as $\{\phi(\mathbf{x}_i), \hat{\mathbf{p}}_i\}_{i=1}^{N}$, which results in slightly smaller training errors while the change of testing errors is not obvious in our experiments on H36M.  
\begin{algorithm}[h]
	\footnotesize
	\caption{Cascaded Deep Networks Training} 
	\hspace*{0.02in} {\bf Input:} \\
	\hspace*{0.02in} Training set $\{\phi(\mathbf{x}_i),\mathbf{p}_i\}_{i=1}^{N}$,
	\hspace*{0.02in} cascade length $T$\\
	\hspace*{0.02in} {\bf Output:} 
	$\mathcal{G}(\mathbf{c}) = \sum_{t=1}^{T}\mathcal{D}_{t}(\mathbf{i}_{t}, \mathbf{\Theta}_{t})$
	\begin{algorithmic}[1]
		\State Current estimate $\{\hat{\mathbf{p}}_i\}_{i=1}^{N} = \mathbf{0}$
		\State Cascade $\mathcal{G}(\mathbf{c}) = 0$
		\For {t=1:$T$}
		\State Inputs $\mathbf{i}_{t} = \{\phi(\mathbf{x}_i)\}_{i=1}^{N}$
		\State Regression targets $= \{\mathbf{p}_i - \hat{\mathbf{p}}_i\}_{i=1}^{N}$
		\State $\mathcal{D}_t$ = $TrainNetwork$(Inputs, Regression targets)
		\State $\mathcal{G}(\mathbf{c}) = \mathcal{G}(\mathbf{c}) + \mathcal{D}_t$
		\For {i=1:N}
		\State $\hat{\mathbf{p}}_i = \hat{\mathbf{p}}_i + \mathcal{D}_t(\phi(\mathbf{x}_i))$
		\EndFor 
		\EndFor
		\State \Return $\mathcal{G}(\mathbf{c})$
	\end{algorithmic}
	\label{alg:cascade}
\end{algorithm}

\subsection{Data Pre-processing}
To train the heatmap regression model $\mathcal{A}(\mathbf{x})$, we download training videos from the \href{http://vision.imar.ro/human3.6m/description.php}{official website} of H36M. We crop the persons with the provided bounding boxes and pad the cropped images with zeros in order to fix the aspect ratio as 4:3. We then resize the padded images to 384 by 288. The target heatmaps have the same size as the input images and we draw a Gaussian dot for each human key-point $\phi(\mathbf{x})_{i=1}^{k}$. The Gaussian dot's mean is the ground truth 2D location of the key-point and it has a standard deviation of 8 pixels. 

To train the cascaded 3D pose estimation model $\mathcal{G}(\mathbf{c})$ on H36M, we download the pre-processed human skeletons released by the authors of \cite{martinez2017simple} in \href{https://github.com/una-dinosauria/3d-pose-baseline}{their github repository}. Each deep learner in the cascaded model is trained with L2 loss. The evaluation set MPI-INF-3DHP is downloaded from the \href{http://gvv.mpi-inf.mpg.de/3dhp-dataset/}{official website} and we use the provided 2D key-points as inputs to evaluate the trained cascaded 2D-to-3D model, which is consistent with and comparable to recent works \cite{Li_2019_CVPR, Ci_2019_ICCV}. 
\section{Ablation Study on Data Evolution} \label{sec3}
\subsection{Effect of Number of Generation $G$} \label{sec3.1}
\begin{figure}
	\centering
	\includegraphics[scale=0.25,  trim=2cm 1cm 2cm 2cm]{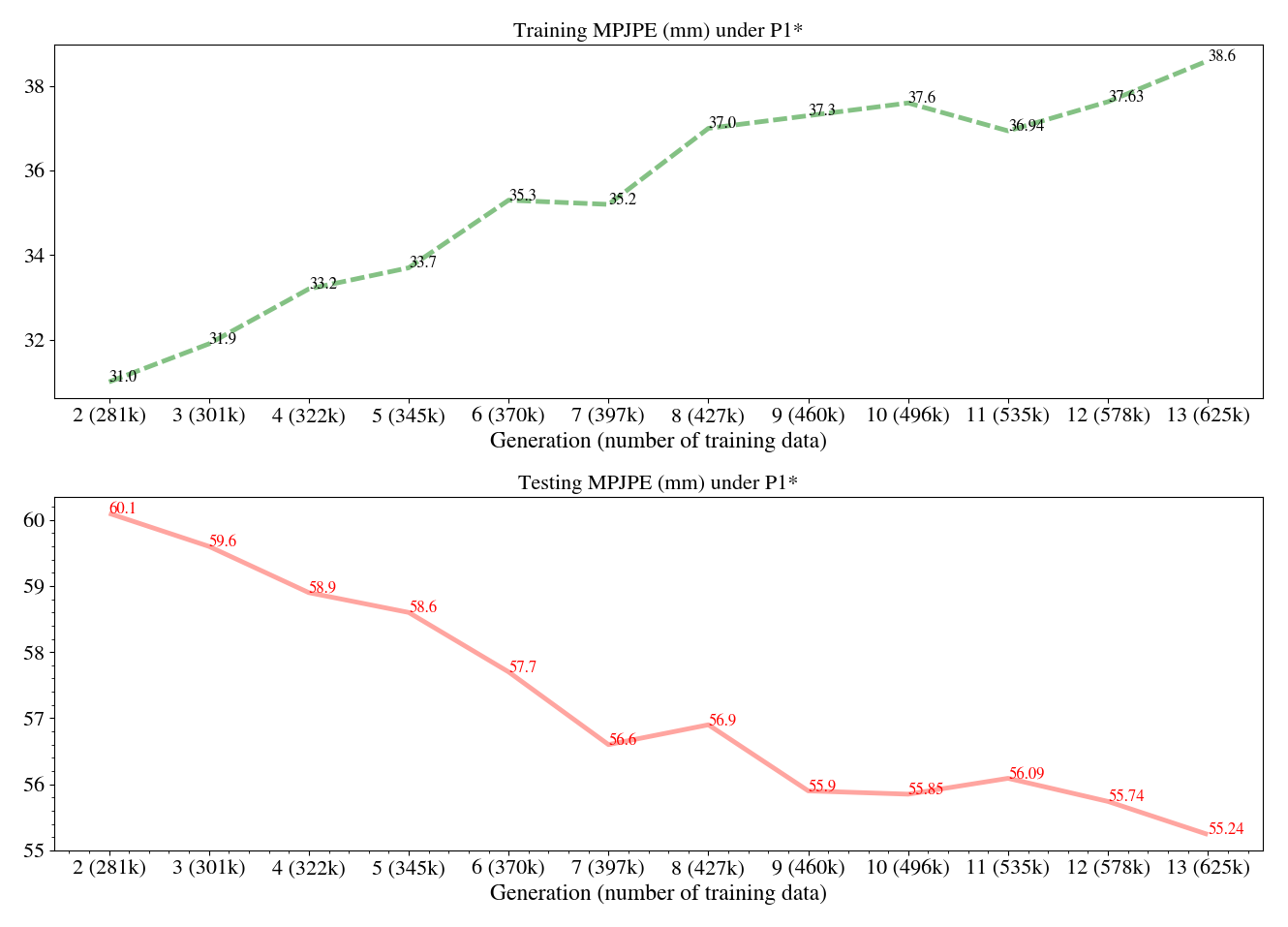}
	\caption{Training and testing MPJPE under P1* with varying number of generation (amount of training 2D-3D pairs).}
	\label{generation}
\end{figure}

To study how the model performance improves with increasing number of synthetic data, we start with S1 data in Human 3.6M and synthesize training datasets with different size by varying the number of generations $G$. We train one model for each evolved dataset. All models have the save architecture and in this study where we fix $T$=1 and $R$=2. These models' performance (MPJPE under P1*) are shown in Fig.~\ref{generation}. Number of generations and the corresponding number of 2D-3D pairs are indicated on the x-axis by (G, N). We observe that the testing errors decrease steadily with larger $G$ (more synthetic data) while the training errors have the opposite trend. This indicates that our data evolution approach indeed synthesizes novel supervision to augment the original dataset. The changing performance of the model can be seen as it is \emph{evolving} along with the growing training dataset, where the model generalization power is significantly improved.  
\subsection{Effect of Evolutionary Operators}
\begin{figure}
	\centering
	\includegraphics[scale=0.3, trim=2cm 1cm 2cm 2cm]{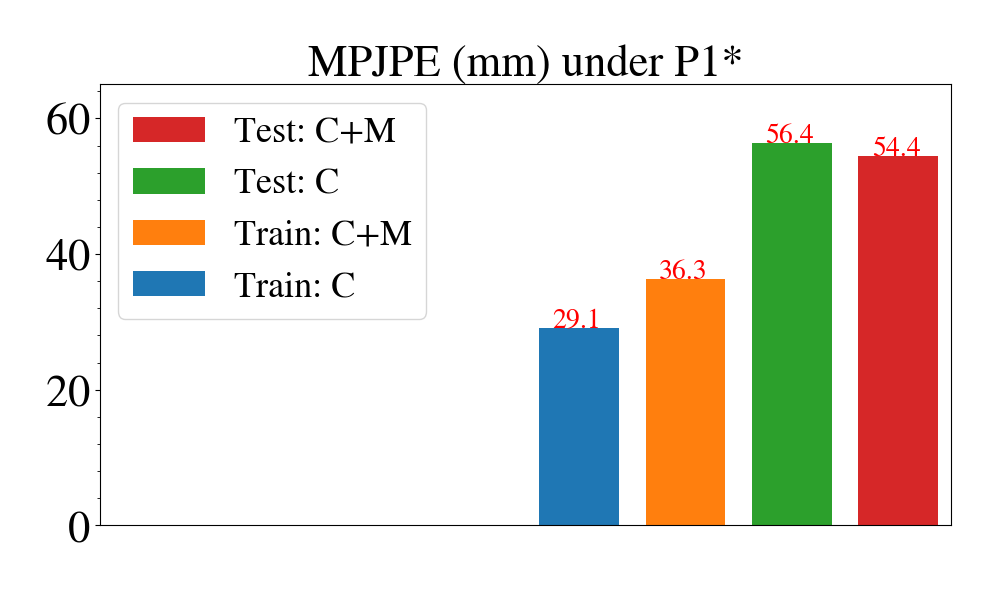}
	\caption{Training and testing errors under P1* with and without adding mutation. C: crossover, M: mutation. Using crossover and mutation together out-performs using crossover alone in our experiments.}
	\label{cm}
\end{figure}

There has been debates on the individual function of crossover and mutation operators~\cite{spears1993crossover}. In \cite{spears1993crossover} the author shows theoretically that the mutation operator has stronger \emph{disruptive} property while the crossover operator has stronger \emph{constructive} property. Here we conduct empirical experiments to study the effectiveness of these evolutionary operators applied on our problem. Here we compare between data evolution with crossover along  and using both operators. The same initial population and model hyper-parameters as in Section \ref{sec3.1} are adopted. The training and testing MPJPEs are shown in Fig.~\ref{cm}. We observe that adding mutation (+M) slightly increases training errors but decreases testing errors just like adding more data in Section \ref{sec3.1}. Despite the difference is not huge, this indicates that using both operators is beneficial for our problem.

\section{Details for U3DPW} \label{sec4}
\begin{figure*}[t]
	\centering
	\includegraphics[scale=0.50, trim=2cm 0cm 3cm 2cm]{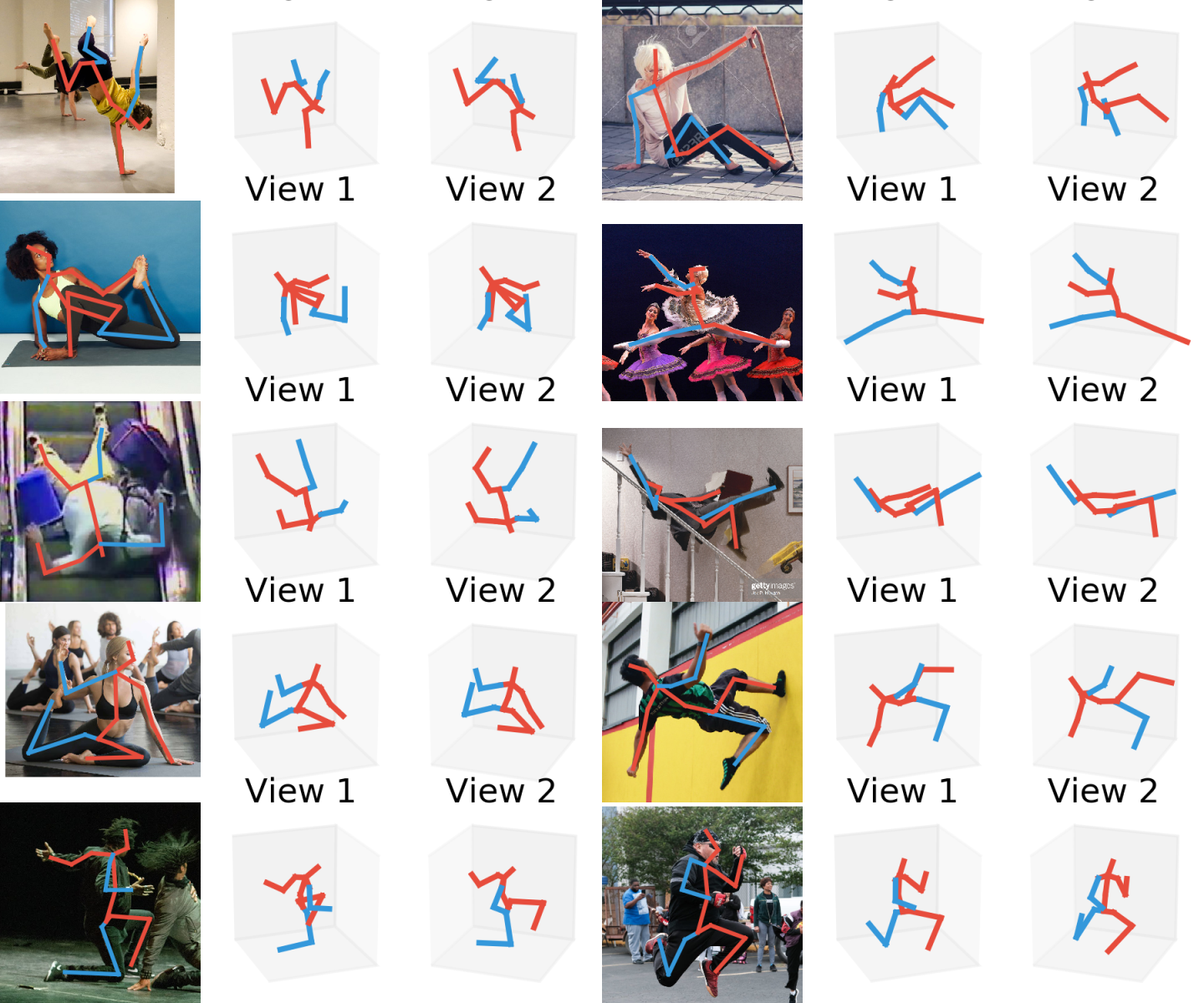}
	\caption{3D annotation examples for U3DPW shown in two different viewpoints.}
	\label{annot3d}
\end{figure*}

\begin{figure}
	\centering
	\includegraphics[scale=0.15]{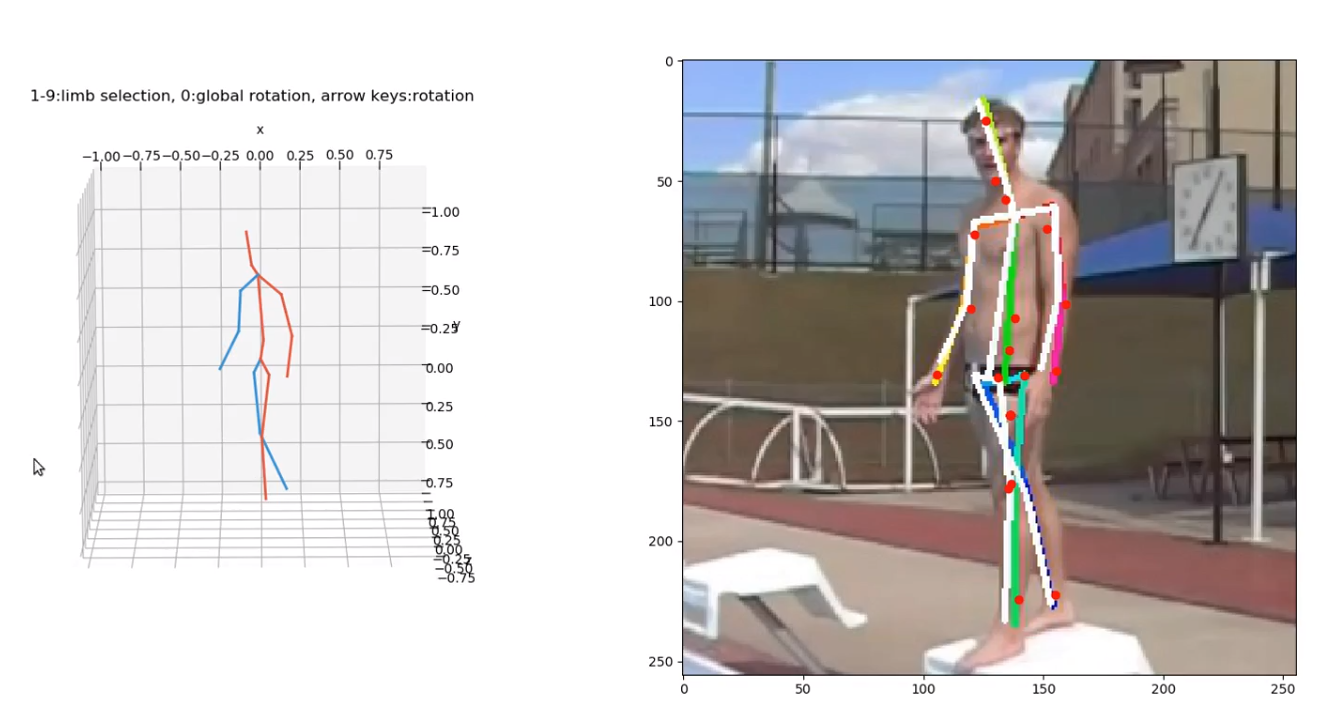}
	\caption{A snapshot of the annotation process. Left: 3D human skeleton. Right: image and 2D projections.}
	\label{tool}
\end{figure}
In this section, we describe how we collect and annotate the images for our new dataset U3DPW.
\subsection{Image Selection}
We start by collecting 300 in-the-wild images that contain humans whose pose is not constrained to daily actions. We select 150 images from the existing LSP dataset \cite{Johnson10} and the remaining 150 high-resolution images are gathered from the Internet. To choose from the LSP dataset, we run SMPLify~\cite{bogo2016keep} on all available images and manually select 150 images with large fitting errors.    

\subsection{2D Annotation}
\begin{figure*}
	\centering
	\includegraphics[scale=1]{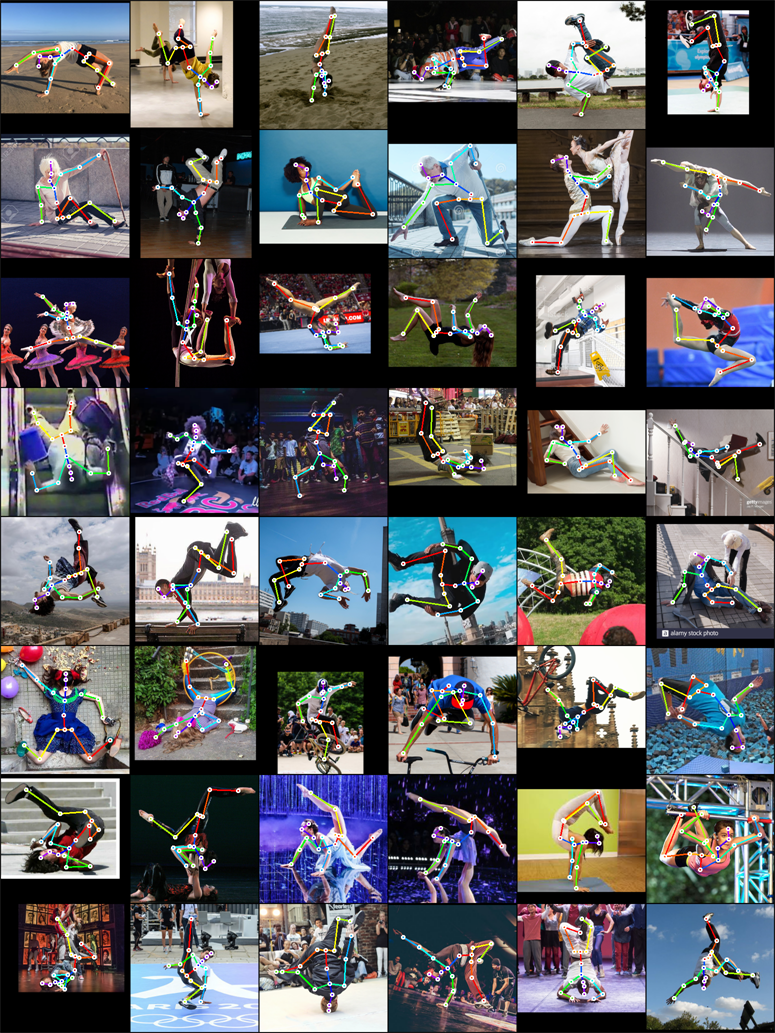}
	\caption{Exemplar images of U3DPW with 2D annotations.}
	\label{batch1}
\end{figure*}

We annotate 17 semantic 2D key-points for each image in U3DPW. These key-points are: right ankle, right knee, right hip, left hip, left knee, left ankle, right wrist, right elbow, right shoulder, left shoulder, left elbow, left wrist, neck, head top, spine, thorax and nose. Example new images (not in LSP) with 2D annotations are shown in Fig.~\ref{batch1}. These images include large variation of human poses, camera viewpoints and illumination. Although we focus on 3D pose estimation in this work, these 2D annotations can also be used to evaluate 2D pose estimation models for unconstrained scenarios.

\subsection{3D Annotation}
Using a subset of 2D key-points we run SMPLify~\cite{bogo2016keep} to obtain fitted 3D human poses. 

We provide an annotation tools that can be used after running SMPLify. The annotation tool displays the 3D skeleton after SMPLify fitting, current image and the projected 2D key-points (calculated with the camera intrinsics from the SMPLify fitting results). The fitted skeleton is converted into a hierarchical human representation, and the user can interactively modify the global pose orientation and the local bone vector orientation by keyboard inputs. With the user feeding inputs, new positions of the 2D human key-points are updated in real-time so that the user can align the 2D projections with some reference key-points. One video demo of operating the annotation tool is placed at \url{root/videos/annotation\_tool.mkv}. Here \url{root} refers to the directory of the unzipped supplementary material folder. Some exemplar 3D annotations are shown in Fig.~\ref{annot3d}.  

\section{Others} \label{sec5}
\subsection{Video for Qualitative Comparison}
The qualitative comparison with \cite{Li_2019_CVPR} (Fig.~\ref{qualitative_comp}) is included as video for better visualization. The file can be found at \url{root/videos/annotation/qualitative.mkv}
\subsection{Failure Cases}
Some individual cases are included as videos in \url{root/videos/}. Projection ambiguity is hard to resolve in some cases and image features should be incorporated instead of only using key-points as inputs.

\clearpage
{\small
	\bibliographystyle{ieee_fullname}
	\bibliography{egbib}

\begin{thebibliography}{10}\itemsep=-1pt

\bibitem{agarwal2005recovering}
Ankur Agarwal and Bill Triggs.
\newblock Recovering 3d human pose from monocular images.
\newblock {\em IEEE transactions on pattern analysis and machine intelligence},
  28(1):44--58, 2005.

\bibitem{akhter2015pose}
Ijaz Akhter and Michael~J Black.
\newblock Pose-conditioned joint angle limits for 3d human pose reconstruction.
\newblock In {\em Proceedings of the IEEE conference on computer vision and
  pattern recognition}, pages 1446--1455, 2015.

\bibitem{andriluka14cvpr}
Mykhaylo Andriluka, Leonid Pishchulin, Peter Gehler, and Bernt Schiele.
\newblock 2d human pose estimation: New benchmark and state of the art
  analysis.
\newblock In {\em IEEE Conference on Computer Vision and Pattern Recognition
  (CVPR)}, June 2014.

\bibitem{anguelov2005scape}
Dragomir Anguelov, Praveen Srinivasan, Daphne Koller, Sebastian Thrun, Jim
  Rodgers, and James Davis.
\newblock Scape: shape completion and animation of people.
\newblock In {\em ACM transactions on graphics (TOG)}, volume~24, pages
  408--416. ACM, 2005.

\bibitem{belagiannis20143d}
Vasileios Belagiannis, Sikandar Amin, Mykhaylo Andriluka, Bernt Schiele, Nassir
  Navab, and Slobodan Ilic.
\newblock 3d pictorial structures for multiple human pose estimation.
\newblock In {\em Proceedings of the IEEE Conference on Computer Vision and
  Pattern Recognition}, pages 1669--1676, 2014.

\bibitem{bo2009structured}
Liefeng Bo and Cristian Sminchisescu.
\newblock Structured output-associative regression.
\newblock In {\em 2009 IEEE Conference on Computer Vision and Pattern
  Recognition}, pages 2403--2410. IEEE, 2009.

\bibitem{bogo2016keep}
Federica Bogo, Angjoo Kanazawa, Christoph Lassner, Peter Gehler, Javier Romero,
  and Michael~J Black.
\newblock Keep it smpl: Automatic estimation of 3d human pose and shape from a
  single image.
\newblock In {\em European Conference on Computer Vision}, pages 561--578.
  Springer, 2016.

\bibitem{burenius20133d}
Magnus Burenius, Josephine Sullivan, and Stefan Carlsson.
\newblock 3d pictorial structures for multiple view articulated pose
  estimation.
\newblock In {\em Proceedings of the IEEE Conference on Computer Vision and
  Pattern Recognition}, pages 3618--3625, 2013.

\bibitem{cai2019exploiting}
Yujun Cai, Liuhao Ge, Jun Liu, Jianfei Cai, Tat-Jen Cham, Junsong Yuan, and
  Nadia~Magnenat Thalmann.
\newblock Exploiting spatial-temporal relationships for 3d pose estimation via
  graph convolutional networks.
\newblock In {\em Proceedings of the IEEE/CVF International Conference on
  Computer Vision}, pages 2272--2281, 2019.

\bibitem{cao2014face}
Xudong Cao, Yichen Wei, Fang Wen, and Jian Sun.
\newblock Face alignment by explicit shape regression.
\newblock {\em International Journal of Computer Vision}, 107(2):177--190,
  2014.

\bibitem{chen2016synthesizing}
Wenzheng Chen, Huan Wang, Yangyan Li, Hao Su, Zhenhua Wang, Changhe Tu, Dani
  Lischinski, Daniel Cohen-Or, and Baoquan Chen.
\newblock Synthesizing training images for boosting human 3d pose estimation.
\newblock In {\em 2016 Fourth International Conference on 3D Vision (3DV)},
  pages 479--488. IEEE, 2016.

\bibitem{chen2019weakly}
Xipeng Chen, Kwan-Yee Lin, Wentao Liu, Chen Qian, and Liang Lin.
\newblock Weakly-supervised discovery of geometry-aware representation for 3d
  human pose estimation.
\newblock In {\em Proceedings of the IEEE Conference on Computer Vision and
  Pattern Recognition}, pages 10895--10904, 2019.

\bibitem{chen2018cascaded}
Yilun Chen, Zhicheng Wang, Yuxiang Peng, Zhiqiang Zhang, Gang Yu, and Jian Sun.
\newblock Cascaded pyramid network for multi-person pose estimation.
\newblock In {\em Proceedings of the IEEE Conference on Computer Vision and
  Pattern Recognition}, pages 7103--7112, 2018.

\bibitem{Cheng_2019_ICCV}
Yu Cheng, Bo Yang, Bo Wang, Wending Yan, and Robby~T. Tan.
\newblock Occlusion-aware networks for 3d human pose estimation in video.
\newblock In {\em The IEEE International Conference on Computer Vision (ICCV)},
  October 2019.

\bibitem{Ci_2019_ICCV}
Hai Ci, Chunyu Wang, Xiaoxuan Ma, and Yizhou Wang.
\newblock Optimizing network structure for 3d human pose estimation.
\newblock In {\em The IEEE International Conference on Computer Vision (ICCV)},
  October 2019.

\bibitem{correia2019evolutionary}
Jo{\~a}o Correia, Tiago Martins, and Penousal Machado.
\newblock Evolutionary data augmentation in deep face detection.
\newblock In {\em Proceedings of the Genetic and Evolutionary Computation
  Conference Companion}, pages 163--164, 2019.

\bibitem{fang2018learning}
Hao-Shu Fang, Yuanlu Xu, Wenguan Wang, Xiaobai Liu, and Song-Chun Zhu.
\newblock Learning pose grammar to encode human body configuration for 3d pose
  estimation.
\newblock In {\em Thirty-Second AAAI Conference on Artificial Intelligence},
  2018.

\bibitem{habibie2019wild}
Ikhsanul Habibie, Weipeng Xu, Dushyant Mehta, Gerard Pons-Moll, and Christian
  Theobalt.
\newblock In the wild human pose estimation using explicit 2d features and
  intermediate 3d representations.
\newblock In {\em Proceedings of the IEEE Conference on Computer Vision and
  Pattern Recognition}, pages 10905--10914, 2019.

\bibitem{he2016deep}
Kaiming He, Xiangyu Zhang, Shaoqing Ren, and Jian Sun.
\newblock Deep residual learning for image recognition.
\newblock In {\em Proceedings of the IEEE conference on computer vision and
  pattern recognition}, pages 770--778, 2016.

\bibitem{holland1992adaptation}
John~Henry Holland et~al.
\newblock {\em Adaptation in natural and artificial systems: an introductory
  analysis with applications to biology, control, and artificial intelligence}.
\newblock MIT press, 1992.

\bibitem{ioffe2015batch}
Sergey Ioffe and Christian Szegedy.
\newblock Batch normalization: Accelerating deep network training by reducing
  internal covariate shift.
\newblock {\em arXiv preprint arXiv:1502.03167}, 2015.

\bibitem{ionescu2013human3}
Catalin Ionescu, Dragos Papava, Vlad Olaru, and Cristian Sminchisescu.
\newblock Human3. 6m: Large scale datasets and predictive methods for 3d human
  sensing in natural environments.
\newblock {\em IEEE transactions on pattern analysis and machine intelligence},
  36(7):1325--1339, 2013.

\bibitem{Johnson10}
Sam Johnson and Mark Everingham.
\newblock Clustered pose and nonlinear appearance models for human pose
  estimation.
\newblock In {\em Proceedings of the British Machine Vision Conference}, 2010.
\newblock doi:10.5244/C.24.12.

\bibitem{kanazawa2018end}
Angjoo Kanazawa, Michael~J Black, David~W Jacobs, and Jitendra Malik.
\newblock End-to-end recovery of human shape and pose.
\newblock In {\em Proceedings of the IEEE Conference on Computer Vision and
  Pattern Recognition}, pages 7122--7131, 2018.

\bibitem{kocabas2019self}
Muhammed Kocabas, Salih Karagoz, and Emre Akbas.
\newblock Self-supervised learning of 3d human pose using multi-view geometry.
\newblock In {\em Proceedings of the IEEE Conference on Computer Vision and
  Pattern Recognition}, pages 1077--1086, 2019.

\bibitem{Kolotouros_2019_ICCV}
Nikos Kolotouros, Georgios Pavlakos, Michael~J. Black, and Kostas Daniilidis.
\newblock Learning to reconstruct 3d human pose and shape via model-fitting in
  the loop.
\newblock In {\em The IEEE International Conference on Computer Vision (ICCV)},
  October 2019.

\bibitem{lee2018propagating}
Kyoungoh Lee, Inwoong Lee, and Sanghoon Lee.
\newblock Propagating lstm: 3d pose estimation based on joint interdependency.
\newblock In {\em Proceedings of the European Conference on Computer Vision
  (ECCV)}, pages 119--135, 2018.

\bibitem{Li_2019_CVPR}
Chen Li and Gim~Hee Lee.
\newblock Generating multiple hypotheses for 3d human pose estimation with
  mixture density network.
\newblock In {\em Proceedings of the IEEE Conference on Computer Vision and
  Pattern Recognition}, pages 9887--9895, 2019.

\bibitem{li2019repair}
Yi Li and Nuno Vasconcelos.
\newblock Repair: Removing representation bias by dataset resampling.
\newblock In {\em Proceedings of the IEEE Conference on Computer Vision and
  Pattern Recognition}, pages 9572--9581, 2019.

\bibitem{Li_2019_ICCV}
Zhi Li, Xuan Wang, Fei Wang, and Peilin Jiang.
\newblock On boosting single-frame 3d human pose estimation via monocular
  videos.
\newblock In {\em The IEEE International Conference on Computer Vision (ICCV)},
  October 2019.

\bibitem{lin2017recurrent}
Mude Lin, Liang Lin, Xiaodan Liang, Keze Wang, and Hui Cheng.
\newblock Recurrent 3d pose sequence machines.
\newblock In {\em Proceedings of the IEEE Conference on Computer Vision and
  Pattern Recognition}, pages 810--819, 2017.

\bibitem{lin2014microsoft}
Tsung-Yi Lin, Michael Maire, Serge Belongie, James Hays, Pietro Perona, Deva
  Ramanan, Piotr Doll{\'a}r, and C~Lawrence Zitnick.
\newblock Microsoft coco: Common objects in context.
\newblock In {\em European conference on computer vision}, pages 740--755.
  Springer, 2014.

\bibitem{liu2020comprehensive}
Kenkun Liu, Rongqi Ding, Zhiming Zou, Le Wang, and Wei Tang.
\newblock A comprehensive study of weight sharing in graph networks for 3d
  human pose estimation.
\newblock In {\em European Conference on Computer Vision}, pages 318--334.
  Springer, 2020.

\bibitem{loper2015smpl}
Matthew Loper, Naureen Mahmood, Javier Romero, Gerard Pons-Moll, and Michael~J
  Black.
\newblock Smpl: A skinned multi-person linear model.
\newblock {\em ACM transactions on graphics (TOG)}, 34(6):248, 2015.

\bibitem{luo2018orinet}
Chenxu Luo, Xiao Chu, and Alan Yuille.
\newblock Orinet: A fully convolutional network for 3d human pose estimation.
\newblock {\em arXiv preprint arXiv:1811.04989}, 2018.

\bibitem{Luvizon_2018_CVPR}
Diogo~C Luvizon, David Picard, and Hedi Tabia.
\newblock 2d/3d pose estimation and action recognition using multitask deep
  learning.
\newblock In {\em Proceedings of the IEEE Conference on Computer Vision and
  Pattern Recognition}, pages 5137--5146, 2018.

\bibitem{martinez2017simple}
Julieta Martinez, Rayat Hossain, Javier Romero, and James~J Little.
\newblock A simple yet effective baseline for 3d human pose estimation.
\newblock In {\em Proceedings of the IEEE International Conference on Computer
  Vision}, pages 2640--2649, 2017.

\bibitem{mehta2017monocular}
Dushyant Mehta, Helge Rhodin, Dan Casas, Pascal Fua, Oleksandr Sotnychenko,
  Weipeng Xu, and Christian Theobalt.
\newblock Monocular 3d human pose estimation in the wild using improved cnn
  supervision.
\newblock In {\em 2017 International Conference on 3D Vision (3DV)}, pages
  506--516. IEEE, 2017.

\bibitem{mehta2018single}
Dushyant Mehta, Oleksandr Sotnychenko, Franziska Mueller, Weipeng Xu, Srinath
  Sridhar, Gerard Pons-Moll, and Christian Theobalt.
\newblock Single-shot multi-person 3d pose estimation from monocular rgb.
\newblock In {\em 2018 International Conference on 3D Vision (3DV)}, pages
  120--130. IEEE, 2018.

\bibitem{mehta2017vnect}
Dushyant Mehta, Srinath Sridhar, Oleksandr Sotnychenko, Helge Rhodin, Mohammad
  Shafiei, Hans-Peter Seidel, Weipeng Xu, Dan Casas, and Christian Theobalt.
\newblock Vnect: Real-time 3d human pose estimation with a single rgb camera.
\newblock {\em ACM Transactions on Graphics (TOG)}, 36(4):44, 2017.

\bibitem{Moon_2019_ICCV_3DMPPE}
Gyeongsik Moon, Juyong Chang, and Kyoung~Mu Lee.
\newblock Camera distance-aware top-down approach for 3d multi-person pose
  estimation from a single rgb image.
\newblock In {\em The IEEE Conference on International Conference on Computer
  Vision (ICCV)}, 2019.

\bibitem{moreno20173d}
Francesc Moreno-Noguer.
\newblock 3d human pose estimation from a single image via distance matrix
  regression.
\newblock In {\em Proceedings of the IEEE Conference on Computer Vision and
  Pattern Recognition}, pages 2823--2832, 2017.

\bibitem{nibali2018numerical}
Aiden Nibali, Zhen He, Stuart Morgan, and Luke Prendergast.
\newblock Numerical coordinate regression with convolutional neural networks.
\newblock {\em arXiv preprint arXiv:1801.07372}, 2018.

\bibitem{nie2017monocular}
Bruce~Xiaohan Nie, Ping Wei, and Song-Chun Zhu.
\newblock Monocular 3d human pose estimation by predicting depth on joints.
\newblock In {\em 2017 IEEE International Conference on Computer Vision
  (ICCV)}, pages 3467--3475. IEEE, 2017.

\bibitem{SMPL-X:2019}
Georgios Pavlakos, Vasileios Choutas, Nima Ghorbani, Timo Bolkart, Ahmed~AA
  Osman, Dimitrios Tzionas, and Michael~J Black.
\newblock Expressive body capture: 3d hands, face, and body from a single
  image.
\newblock In {\em Proceedings of the IEEE Conference on Computer Vision and
  Pattern Recognition}, pages 10975--10985, 2019.

\bibitem{pavlakos2018ordinal}
Georgios Pavlakos, Xiaowei Zhou, and Kostas Daniilidis.
\newblock Ordinal depth supervision for 3d human pose estimation.
\newblock In {\em Proceedings of the IEEE Conference on Computer Vision and
  Pattern Recognition}, pages 7307--7316, 2018.

\bibitem{pavlakos2017coarse}
Georgios Pavlakos, Xiaowei Zhou, Konstantinos~G Derpanis, and Kostas
  Daniilidis.
\newblock Coarse-to-fine volumetric prediction for single-image 3d human pose.
\newblock In {\em Proceedings of the IEEE Conference on Computer Vision and
  Pattern Recognition}, pages 7025--7034, 2017.

\bibitem{Pavlakos_2017_CVPR}
Georgios Pavlakos, Xiaowei Zhou, Konstantinos~G Derpanis, and Kostas
  Daniilidis.
\newblock Harvesting multiple views for marker-less 3d human pose annotations.
\newblock In {\em Proceedings of the IEEE conference on computer vision and
  pattern recognition}, pages 6988--6997, 2017.

\bibitem{Pavllo_2019_CVPR}
Dario Pavllo, Christoph Feichtenhofer, David Grangier, and Michael Auli.
\newblock 3d human pose estimation in video with temporal convolutions and
  semi-supervised training.
\newblock In {\em Proceedings of the IEEE Conference on Computer Vision and
  Pattern Recognition}, pages 7753--7762, 2019.

\bibitem{peng2018jointly}
Xi Peng, Zhiqiang Tang, Fei Yang, Rogerio~S Feris, and Dimitris Metaxas.
\newblock Jointly optimize data augmentation and network training: Adversarial
  data augmentation in human pose estimation.
\newblock In {\em Proceedings of the IEEE Conference on Computer Vision and
  Pattern Recognition}, pages 2226--2234, 2018.

\bibitem{Hossain_2018_ECCV}
Mir Rayat Imtiaz~Hossain and James~J Little.
\newblock Exploiting temporal information for 3d human pose estimation.
\newblock In {\em Proceedings of the European Conference on Computer Vision
  (ECCV)}, pages 68--84, 2018.

\bibitem{Ren_2014_CVPR}
Shaoqing Ren, Xudong Cao, Yichen Wei, and Jian Sun.
\newblock Face alignment at 3000 fps via regressing local binary features.
\newblock In {\em Proceedings of the IEEE Conference on Computer Vision and
  Pattern Recognition}, pages 1685--1692, 2014.

\bibitem{Rhodin_2018_ECCV}
Helge Rhodin, Mathieu Salzmann, and Pascal Fua.
\newblock Unsupervised geometry-aware representation for 3d human pose
  estimation.
\newblock In {\em Proceedings of the European Conference on Computer Vision
  (ECCV)}, pages 750--767, 2018.

\bibitem{Rhodin_2018_CVPR}
Helge Rhodin, J{\"o}rg Sp{\"o}rri, Isinsu Katircioglu, Victor Constantin,
  Fr{\'e}d{\'e}ric Meyer, Erich M{\"u}ller, Mathieu Salzmann, and Pascal Fua.
\newblock Learning monocular 3d human pose estimation from multi-view images.
\newblock In {\em Proceedings of the IEEE Conference on Computer Vision and
  Pattern Recognition}, pages 8437--8446, 2018.

\bibitem{rogez2016mocap}
Gr{\'e}gory Rogez and Cordelia Schmid.
\newblock Mocap-guided data augmentation for 3d pose estimation in the wild.
\newblock In {\em Advances in neural information processing systems}, pages
  3108--3116, 2016.

\bibitem{rogez2017lcr}
Gregory Rogez, Philippe Weinzaepfel, and Cordelia Schmid.
\newblock Lcr-net: Localization-classification-regression for human pose.
\newblock In {\em Proceedings of the IEEE Conference on Computer Vision and
  Pattern Recognition}, pages 3433--3441, 2017.

\bibitem{rosales2002learning}
R{\'o}mer Rosales and Stan Sclaroff.
\newblock Learning body pose via specialized maps.
\newblock In {\em Advances in neural information processing systems}, pages
  1263--1270, 2002.

\bibitem{Sharma_2019_ICCV}
Saurabh Sharma, Pavan~Teja Varigonda, Prashast Bindal, Abhishek Sharma, and
  Arjun Jain.
\newblock Monocular 3d human pose estimation by generation and ordinal ranking.
\newblock In {\em The IEEE International Conference on Computer Vision (ICCV)},
  October 2019.

\bibitem{sharma2019monocular}
Saurabh Sharma, Pavan~Teja Varigonda, Prashast Bindal, Abhishek Sharma, and
  Arjun Jain.
\newblock Monocular 3d human pose estimation by generation and ordinal ranking.
\newblock In {\em Proceedings of the IEEE/CVF International Conference on
  Computer Vision}, pages 2325--2334, 2019.

\bibitem{shi2016real}
Wenzhe Shi, Jose Caballero, Ferenc Husz{\'a}r, Johannes Totz, Andrew~P Aitken,
  Rob Bishop, Daniel Rueckert, and Zehan Wang.
\newblock Real-time single image and video super-resolution using an efficient
  sub-pixel convolutional neural network.
\newblock In {\em Proceedings of the IEEE conference on computer vision and
  pattern recognition}, pages 1874--1883, 2016.

\bibitem{sigal2010humaneva}
Leonid Sigal, Alexandru~O Balan, and Michael~J Black.
\newblock Humaneva: Synchronized video and motion capture dataset and baseline
  algorithm for evaluation of articulated human motion.
\newblock {\em International journal of computer vision}, 87(1-2):4, 2010.

\bibitem{spears1993crossover}
William~M Spears.
\newblock Crossover or mutation?
\newblock In {\em Foundations of genetic algorithms}, volume~2, pages 221--237.
  Elsevier, 1993.

\bibitem{srivastava2014dropout}
Nitish Srivastava, Geoffrey Hinton, Alex Krizhevsky, Ilya Sutskever, and Ruslan
  Salakhutdinov.
\newblock Dropout: a simple way to prevent neural networks from overfitting.
\newblock {\em The journal of machine learning research}, 15(1):1929--1958,
  2014.

\bibitem{sun2019deep}
Ke Sun, Bin Xiao, Dong Liu, and Jingdong Wang.
\newblock Deep high-resolution representation learning for human pose
  estimation.
\newblock In {\em Proceedings of the IEEE Conference on Computer Vision and
  Pattern Recognition}, pages 5693--5703, 2019.

\bibitem{sun_2018_eccv}
Xiao Sun, Bin Xiao, Fangyin Wei, Shuang Liang, and Yichen Wei.
\newblock Integral human pose regression.
\newblock In {\em Proceedings of the European Conference on Computer Vision
  (ECCV)}, pages 529--545, 2018.

\bibitem{tommasi2017deeper}
Tatiana Tommasi, Novi Patricia, Barbara Caputo, and Tinne Tuytelaars.
\newblock A deeper look at dataset bias.
\newblock In {\em Domain adaptation in computer vision applications}, pages
  37--55. Springer, 2017.

\bibitem{torralba2011unbiased}
A Torralba and AA Efros.
\newblock Unbiased look at dataset bias.
\newblock In {\em Proceedings of the 2011 IEEE Conference on Computer Vision
  and Pattern Recognition}, pages 1521--1528. IEEE Computer Society, 2011.

\bibitem{varol2017learning}
Gul Varol, Javier Romero, Xavier Martin, Naureen Mahmood, Michael~J Black, Ivan
  Laptev, and Cordelia Schmid.
\newblock Learning from synthetic humans.
\newblock In {\em Proceedings of the IEEE Conference on Computer Vision and
  Pattern Recognition}, pages 109--117, 2017.

\bibitem{Wandt_2019_CVPR}
Bastian Wandt and Bodo Rosenhahn.
\newblock Repnet: Weakly supervised training of an adversarial reprojection
  network for 3d human pose estimation.
\newblock In {\em The IEEE Conference on Computer Vision and Pattern
  Recognition (CVPR)}, June 2019.

\bibitem{wandt2019repnet}
Bastian Wandt and Bodo Rosenhahn.
\newblock Repnet: Weakly supervised training of an adversarial reprojection
  network for 3d human pose estimation.
\newblock In {\em Proceedings of the IEEE Conference on Computer Vision and
  Pattern Recognition}, pages 7782--7791, 2019.

\bibitem{yang20183d}
Wei Yang, Wanli Ouyang, Xiaolong Wang, Jimmy Ren, Hongsheng Li, and Xiaogang
  Wang.
\newblock 3d human pose estimation in the wild by adversarial learning.
\newblock In {\em Proceedings of the IEEE Conference on Computer Vision and
  Pattern Recognition}, pages 5255--5264, 2018.

\bibitem{zhang2013actemes}
Weiyu Zhang, Menglong Zhu, and Konstantinos~G Derpanis.
\newblock From actemes to action: A strongly-supervised representation for
  detailed action understanding.
\newblock In {\em Proceedings of the IEEE International Conference on Computer
  Vision}, pages 2248--2255, 2013.

\bibitem{Zhao_2019_CVPR}
Long Zhao, Xi Peng, Yu Tian, Mubbasir Kapadia, and Dimitris~N Metaxas.
\newblock Semantic graph convolutional networks for 3d human pose regression.
\newblock In {\em Proceedings of the IEEE Conference on Computer Vision and
  Pattern Recognition}, pages 3425--3435, 2019.

\bibitem{zhou2017towards}
Xingyi Zhou, Qixing Huang, Xiao Sun, Xiangyang Xue, and Yichen Wei.
\newblock Towards 3d human pose estimation in the wild: a weakly-supervised
  approach.
\newblock In {\em Proceedings of the IEEE International Conference on Computer
  Vision}, pages 398--407, 2017.

\bibitem{Zhou_2016_CVPR}
Xiaowei Zhou, Menglong Zhu, Spyridon Leonardos, Konstantinos~G Derpanis, and
  Kostas Daniilidis.
\newblock Sparseness meets deepness: 3d human pose estimation from monocular
  video.
\newblock In {\em Proceedings of the IEEE conference on computer vision and
  pattern recognition}, pages 4966--4975, 2016.

\end{thebibliography}
}
\end{document}